\documentclass{article}

\PassOptionsToPackage{numbers, compress}{natbib}

\usepackage[preprint]{neurips_data_2023}

\usepackage{color}
\usepackage{soul}
\usepackage[dvipsnames]{xcolor}
\usepackage[normalem]{ulem}
\newcommand{\todo}[1]{\textcolor{BrickRed}{[\textbf{TODO}]} }

\usepackage[utf8]{inputenc} 
\usepackage[T1]{fontenc}    
\usepackage{hyperref}       
\usepackage{url}            
\usepackage{booktabs}       
\usepackage{nicefrac}       
\usepackage{microtype}      
\usepackage{xcolor}         
\usepackage{multirow}
\usepackage{graphicx}
\usepackage{amsfonts}       
\usepackage{amsmath}
\usepackage{nicefrac} 
\usepackage{lscape}

\usepackage{listings}
\usepackage[frozencache=true,cachedir=minted-cache]{minted} 

\input{content/custom}

\title{\textsc{OceanBench}: \\ The Sea Surface Height Edition}

\author{%
  J. Emmanuel Johnson$^*$\\
  CNRS UMR IGE \\
  \texttt{johnsonj@univ-grenoble-alpes.fr}\\
  \And
  Quentin Febvre$^*$\\
  IMT Atlantique\\
  \texttt{quentin.febvre@imt-atlantique.fr} \\
  \And
  Anastasia Gorbunova\\
  CNRS UMR IGE \\
  \And
  Sammy Metref\\
  DATLAS\\
  \And
  Maxime Ballarotta\\
  CLS\\
  \And
  Julien Le Sommer\\
  CNRS UMR IGE\\
  \And
  Ronan Fablet\\
  IMT Atlantique \\
}

\begin{document}

\maketitle
\def\thefootnote{*}\footnotetext{These authors contributed equally to this work}\def\thefootnote{\arabic{footnote}}

\begin{abstract}

The ocean is a crucial component of the Earth's system. 
It profoundly influences human activities and plays a critical role in climate regulation. 
Our understanding has significantly improved over the last decades with the advent of satellite remote sensing data, allowing us to capture essential sea surface quantities over the globe, e.g., sea surface height (SSH). 
Despite their ever-increasing abundance, ocean satellite data presents challenges for information extraction due to their sparsity and irregular sampling, signal complexity, and noise. 
Machine learning (ML) techniques have demonstrated their capabilities in dealing with large-scale, complex signals. 
Therefore we see an opportunity for these ML models to harness the full extent of the information contained in ocean satellite data. 
However, data representation and relevant evaluation metrics can be \textit{the} defining factors when determining the success of applied ML. 
The processing steps from the raw observation data to a ML-ready state and from model outputs to interpretable quantities require domain expertise, which can be a significant barrier to entry for ML researchers. 
In addition, imposing fixed processing steps, like committing to specific variables, regions, and geometries, will narrow the scope of ML models and their potential impact on real-world applications. 
\textbf{OceanBench} is a unifying framework that provides standardized processing steps that comply with domain-expert standards. 
It is designed with a flexible and pedagogical abstraction: it a) provides plug-and-play data and pre-configured pipelines for ML researchers to benchmark their models w.r.t. ML and domain-related baselines and b) provides a transparent and configurable framework for researchers to customize and extend the pipeline for their tasks. 
In this work, we demonstrate the \texttt{OceanBench} framework through a first edition dedicated to SSH interpolation challenges. 
We provide datasets and ML-ready benchmarking pipelines for the long-standing problem of interpolating observations from simulated ocean satellite data, multi-modal and multi-sensor fusion issues, and transfer-learning to real ocean satellite observations. 
The \texttt{OceanBench} framework is available at~\href{https://github.com/jejjohnson/oceanbench}{github.com/jejjohnson/oceanbench} and the dataset registry is available at~\href{https://github.com/quentinf00/oceanbench-data-registry}{github.com/quentinf00/oceanbench-data-registry}.

\end{abstract}

\section{Motivation}

The ocean is vital to the Earth's system~\cite{OCEANWARMING}. 
It plays a significant role in climate regulation regarding carbon~\cite{OCEANCARBONCYCLE} and heat uptake~\cite{OCEANHEATUPTAKE}. It is also a primary driver of human activities (e.g., maritime traffic and world trade, marine resources and services)~\cite{SSHOPERATIONAL, ML4OCN}. 
However, monitoring the ocean is a critical challenge: the ocean state can only partially be determined because most of the ocean consists of subsurface quantities that we cannot directly observe. 
Thus, to quantify even a fraction of the physical or biochemical ocean state, we must often rely only on surface quantities that we can monitor from space, drifting buoys, or autonomous devices.
Satellite remote sensing, in particular, is one of the most effective ways of measuring essential sea surface quantities~\cite{Altimetry} such as sea surface height (SSH)~\cite{DUACS}, sea surface temperature (SST)~\cite{OCEANSATELLITESST}, and ocean color (OC)~\cite{OCEANSATELLITEOC}. 
While these variables characterize only a tiny portion of the ocean ecosystem, they present a gateway to many other derived physical quantities~\cite{ML4OCN}.

Although we can access observable sea surface quantities, they are generally irregularly and extremely sparsely sampled. 
For instance, satellite-derived SSH data has less than 5\% coverage of the globe daily~\cite{DUACS}. 
These sampling gaps make the characterization of ocean processes highly challenging for operational products and downstream tasks that depend on relevant gap-free variables. This has motivated a rich literature in geoscience over the last decades, mainly using geostatistical kriging methods \cite{DUACS, MIOST} and model-driven data assimilation schemes~\cite{BFNQG, GLORYS12}. Despite significant progress, these schemes often need to improve their ability to leverage available observation datasets' potential fully. 
This has naturally advocated for exploring data-driven approaches like shallow ML schemes~\cite{DINEOF, DINEOF2, ANALOGDA2, ANALOGDA}. Very recently, deep learning schemes \cite{SSHInterpAttention, SSHInterpConvLSTM, SSHInterpUNet} have become appealing solutions to benefit from existing large-scale observation and simulation datasets and reach significant breakthroughs in the monitoring of upper ocean dynamics from scarcely and irregularly sampled observations. However, the heterogeneity and characteristics of the observation data present major challenges for effectively applying these methods beyond idealized case studies. 
A data source could have different variables, geometries, and noise levels, resulting in many domain-specific preprocessing procedures that can vastly change the solution outcome.
Furthermore, the evaluation procedure of the methods and their effectiveness can be regionally-dependent as the physical phenomena vary in space and time, which adds another layer of complexity in convincing domain scientists of their trustworthiness.
So the entire ML pipeline now requires a unified framework for dealing with heterogeneous data sources, different pre- and post-processing methodologies, and regionally-dependent evaluation procedures.


To address these challenges, we introduce \textbf{OceanBench}, a framework for co-designing machine-learning-driven high-level experiments from ocean observations. 
It consists of an end-to-end framework for piping data from its raw form to an ML-ready state and from model outputs to interpretable quantities. 
We regard \texttt{OceanBench} as a key facilitator for the uptake of MLOPs tools and research~\cite{MLOPS1,MLOPS2} for ocean-related datasets and case studies. This first edition provides datasets and ML-ready benchmarking pipelines for SSH interpolation problems, an essential topic for the space oceanography community, related to ML communities dealing with issues like in-painting~\cite{InPaintingSurvey}, denoising~\cite{DENOISESURVEY,DENOISESURVEY2}, and super-resolution~\cite{SuperResSurvey}. 
We expect \texttt{OceanBench} to facilitate new challenges to the applied machine learning community and contribute to meaningful ocean-relevant breakthroughs.
The remainder of the paper is organized as follows: in \S2, we outline some related work that was inspirational for this work; in \S3, we formally outline \texttt{OceanBench} by highlighting the target audience, code structure, and problem scope; in \S4, we outline the problem formulation of SSH interpolation and provide some insight into different tasks related to SSH interpolation where \texttt{OceanBench} could provide some helpful utility; and in \S5 we give some concluding remarks while also informally inviting other researchers to help fill in the gaps.

\section{Related Work}

Machine learning applied to geosciences is becoming increasingly popular, but there are few examples of transparent pipelines involving observation data. 
After a thorough literature review, we have divided the field into three camps of ML applications that pertain to this work: 1) toy simulation datasets, 2) reanalysis datasets, and 3) observation datasets. 
We outline the literature for each of the three categories below.

\textbf{Toy Simulation Data}. 
One set of benchmarks focuses on learning surrogate models for well-defined but chaotic dynamical systems in the form of ordinary differential equations (ODEs) and partial differential equations (PDEs) and there are freely available code bases which implement different ODEs/PDEs~\citep{CHAOSBENCH,PDEBench,pyQG,JAXCFD,NCARDART,NCARDARTSOFTWARE,VEROS,OCEANANIGANS}.
This is a great testing ground for simple toy problems that better mimic the structures we see in real-world observations. 
Working with simulated data is excellent because it is logistically simple and allows users to test their ideas on toy problems without increasing the complexity when dealing with real-world data.
However, these are ultimately simple physical models that often do not reflect the authentic structures we see in real-world, observed data.

\textbf{Reanalysis Data}. 
This is assimilated data of real observations and model simulations. 
There are a few major platforms that host ocean reanalysis data like the Copernicus Marine Data Store~\citep{MDSOCEANPHYSICS,MDSBIOGEOCHEMICAL,MDSOCEANPHYSICSENS,MDSWAVES}, the Climate Data Store~\citep{CDSREANALYSISSST}, the BRAN2020 Model~\citep{DATABLUELINK}, and the NOAA platform~\citep{DATANCEP}. 
However, to our knowledge, there is no standard ML-specific ocean-related tasks to accompany the data. On the atmospheric side, platforms like \texttt{WeatherBench}~\cite{weatherbench}, \texttt{ClimateBench}~\cite{ClimateBench}, \texttt{ENS10}~\cite{ENS10Bench} were designed to assess short-term and medium-term forecasting using ML techniques with recent success of ML~\cite{GraphCast,FourCastNet}
The clarity of the challenges set by the benchmark suites has inspired the idea of \texttt{OceanBench}, where we directly focus on problems dealing with ocean observation data.

\textbf{Observation Data}. 
These observation datasets (typically sparse) stem from satellite observations that measure surface variables or in-situ measurements that measure quantities within the water column. 
Some major platforms to host data include the Marine Data Store~\citep{MDSALONGTRACK,MDSINSITU}, the Climate Data Store~\citep{CDSOBSSST,CDSOBSSSTENS,CDSOBSOC}, ARGO~\citep{ARGO}, and the SOCAT platform~\citep{SOCAT}.
However, it is more difficult to assess the efficacy of operational ML methods that have been trained only on observation data and, to our knowledge, there is no coherent ML benchmarking system for ocean state estimation.
There has been significant effort by the \textit{Ocean-Data-Challenge} Group\footnote{Ocean Data Challenge group: Freely associated scientist for oceanographic algorithm and product improvements (\href{https://ocean-data-challenges.github.io/}{ocean-data-challenges.github.io})} which provides an extensive suite of datasets and metrics for SSH interpolation.
Their efforts heavily inspired our work, and we hope that \texttt{OceanBench} can build upon their work by adding cohesion and facilitating the ease of use for ML research and providing a high-level framework for providing ML-related data products.

\section{OceanBench} \label{sec:oceanbench_intro}

\subsection{Why OceanBench?} \label{sec:oceanbench_why}

There is a high barrier to entry in working with ocean observations for researchers in applied machine learning as there are many processing steps for both the observation data and the domain-specific evaluation procedures. 
\texttt{OceanBench} aims to lower the barrier to entry cost for ML researchers to make meaningful progress in the field of state prediction. 
We distribute a standardized, transparent, and flexible procedure for defining data and evaluation pipelines for data-intensive geoscience applications. 
Proposed examples and case studies provide a plug-and-play framework to benchmark novel ML schemes w.r.t.  state-of-the-art, domain-specific ML baselines. 
In addition, we adopt a pedagogical abstraction that allows users to customize and extend the pipelines for their specific tasks.
To our knowledge, no framework embeds processing steps for earth observation data in a manner compatible with MLOps abstractions and standards regarding reproducibility and evaluation procedures. 
Ultimately, we aim to facilitate the uptake of ML schemes to address ocean observation challenges and to bring new challenges to the ML community to extend additional ML tools and methods for irregularly-sampled and partially-observed high-dimensional space-time dynamics.
The abstractions proposed here apply beyond ocean sciences and SSH interpolation to other geosciences with similar tasks that intersect with machine learning.


\subsection{Code Structure} \label{sec:code_structure}

\texttt{OceanBench} is lightweight in terms of the core functionality.
We keep the code base simple and focus more on how the user can combine each piece.
We adopt a strict functional style because it is easier to maintain and combine sequential transformations. 
There are five features we would like to highlight about \texttt{OceanBench}: 1) Data availability and version control, 2) an agnostic suite of geoprocessing tools for \texttt{xarray} datasets that were aggregated from different sources,  3) Hydra integration to pipe sequential transformations, 4) a flexible multi-dimensional array generator from \texttt{xarray} datasets that are compatible with common deep learning (DL) frameworks, and 5) a JupyterBook~\cite{JupyterBook} that offers library tutorials and demonstrates use-cases.
In the following section, we highlight these components in more detail.

\textbf{Data Availability}. 
The most important aspect is the public availability of the datasets. 
We aggregate all pre-curated datasets from other sources, e.g. the \textit{Ocean-Data-Challenge}~\cite{DCOSEGULFSSH,DCOSSEGULFSSH}, and organize them to be publicly available from a single source~\footnote{Available at: \href{https://github.com/quentinf00/oceanbench-data-registry}{oceanbench-data-registry.github.com}}. 
We also offer a few derived datasets which can be used for demonstrations and evaluation. 
Data is never static in a pipeline setting, as one can have many derived datasets which stem from numerous preprocessing choices. 
In fact, in research, we often work with derived datasets that have already been through some preliminary preprocessing methods. 
To facilitate the ever-changing nature of data, we use the Data Version Control (\texttt{DVC}) tool~\cite{DVC}, which offers a git-like version control of the datasets.

\textbf{Geoprocessing Tools}. 
The core \texttt{OceanBench} library offers a suite of functions specific to processing geo-centric data. 
While a few particular functionalities vary from domain to domain, many operations are standard, e.g., data variable selections, filtering/smoothing, regridding, coordinate transformations, and standardization. 
We almost work exclusively with the \texttt{xarray}~\cite{XARRAY} framework because it is a coordinate-aware, flexible data structure. 
In addition, the geoscience community has an extensive suite of specialized packages that operate in the \texttt{xarray} framework to accomplish many different tasks. 
Almost all \texttt{OceanBench} toolsets are exclusively within the \texttt{xarray} framework to maintain compatibility with a large suite of tools already available from the community.



\textbf{Hydra Integration}. 
As discussed above, many specific packages accomplish many different tasks. 
However, what needs to be added is the flexibility to mix and match these operations as the users see fit. 
\texttt{Hydra}~\cite{Hydra} provides a configurable way to aggregate and \textit{pipe} many sequential operations together. 
It also maintains readability, robustness, and flexibility through the use of \texttt{.yaml} files which explicitly highlights the function used, the function parameters chosen, and the sequence of operations performed. 
In the ML software stack, \texttt{Hydra} is often used to manage the model, optimizer, and loss configurations which helps the user experiment with different options. 
We apply this same concept in preprocessing, geoprocessing, and evaluation steps, often more important than the model configuration in geoscience-related tasks.  

\texttt{XRPatcher}~\footnote{Available at: \href{https://github.com/jejjohnson/xrpatcher/}{github.com/jejjohnson/xrpatcher}}. 
Every machine learning pipeline will inevitably require moving data from the geo-specific data structure to a multi-dimensional array easily digestible for ML models. 
A rather underrated, yet critical, feature of ML frameworks such as \texttt{PyTorch}~\cite{PYTORCH} (\texttt{Lightning}~\cite{LIGHTNING}) and \texttt{TensorFlow}~\cite{TENSORFLOW} (\texttt{Keras}~\cite{KERAS}) is the abstraction of the dataset, dataloader, datamodules, and data pipelines. 
In applied ML in geosciences, the data pipelines are often more important than the actual model~\cite{DATA4ML}. 
The user can control the \textit{patch}-size and the \textit{stride}-step, which can generate arbitrary coordinate-aware items directly from the \texttt{xarray} data structure. 
In addition, \texttt{XRPatcher} provides a way to reconstruct the fields from an arbitrary patch configuration.
This robust reconstruction step is convenient to extend the ML inference step where one can reconstruct entire fields of arbitrary dimensions beyond the training configuration, e.g., to account for the border effects within the field (see appendix~\ref{sec:xrpatcher}) or to reconstruct quantities in specific regions or globally. 

\textbf{JupyterBook}.
Building a set of tools is relatively straightforward; however, ensuring that it sees a broader adoption across a multi-disciplinary community is much more challenging. 
We invested heavily in showing use cases that appeal to different users with the \texttt{JupyterBook} platform~\cite{JupyterBook}. 
Code with context is imperative for domain and ML experts as we need to explain and justify each component and give many examples of how they can be used in other situations. 
Thus, we have paid special attention to providing an extensive suite of tutorials, and we also highlight use cases for how one can effectively use the tools.

\subsection{Problem Scope} \label{sec:problem_scope}


There are many problems that are of great interest the ocean community~\citep{ML4DA} but we limit the scope to state estimation problems~\citep{DAGEOSCIENCE}. Under this scope, there are research questions that are relevant to operational centers which are responsible for generating the vast majority of global ocean state maps~\citep{MDSOCEANPHYSICS,MDSOCEANPHYSICSENS,MDSBIOGEOCHEMICAL, MDSWAVES} that are subsequently used for many downstream tasks~\citep{ML4OCN}. For example: how can we effectively use heterogeneous observations to predict the ocean state on the sea surface~\citep{BFNQG,NERFSSSH,MIOST,4DVARNETSST,4DVARNETSWOT,OCEANSATELLITESST}; how can we incorporate prior physics knowledge into our predictions of ocean state trajectories~\citep{BFNQG,ML4DA,ML4OCN}; and how can we use the current ocean state at time $T$ to predict the future ocean state at time $T+\tau$~\citep{METNET2,weatherbench,FORECASTSSCGP}.
In the same vain, there are more research questions that are of interest to the academic modeling community. For example: is simulated or reanalysis data more effective for learning ML emulators that replace expensive ocean models~\citep{MLSUBGRID,MLCLOSURE}; what metrics are more effective for assessing our ability to mimic ocean dynamics~\citep{SSTFLOWANOMALY,MLMETRICSINVARIANCE}; and how much model error can we characterize when learning from observations~\citep{MLMODELERR,MLMODELERR2}. 

We have cited many potential applications of how ML can be applied to tackle the state estimation problem. 
However, to our knowledge there is no publicly available, standardized benchmark system that is caters to ML-research standards.
We believe that, irrespective of the questions posed above and the data we access, there are many logistical similarities for each of the problem formulations where we can start to set standards for a subset of tasks like interpolation or forecasting. 
On the front-end, we need a way to select regions, periods, variables, and a valid train-test split (see sec. ~\ref{sec:hydra_recipe_task}). 
On the back-end, we need a way to transform the predictions into more meaningful variables with appropriate metrics for validation (see sec. ~\ref{sec:hydra_geoprocess_task} and ~\ref{sec:hydra_evaluation_task}).
\texttt{OceanBench} was designed to be an agnostic tool that is extensible to the types of datasets, processing techniques and metrics needed for working with a specific class of Ocean-related datasets. 
We strongly feel that a suite like this is the first step in designing task-specific benchmarks within the ocean community that is compatible with ML standards. 
In the remainder of the paper, we will demonstrate how \texttt{OceanBench} can be configured to facilite a ML-ready data challenge involving our first edition to demonstrate \texttt{OceanBench}'s applicability: sea surface height interpolation.

\section{\textit{Sea Surface Height Edition}}\label{sec:interp_challenge}

Sea surface height (SSH) is one of the most critical, observable quantities when determining the ocean state. 
It is widely used to study ocean dynamics and the adverse impact on global climate and human activities~\cite{SSHMESOSCALE}. 
SSH enables us to track phenomena such as currents and eddies~\cite{SSHMESOSCALE,SSHMESOSCALE2,SSHMESOSCALE3}, which leads to a better quantification of the transport of energy, heat, and salt. 
In addition, SSH helps us quantify sea level rise at regional and global scales~\cite{SSHSEALEVEL,OCEANSEALEVEL}, which is used for operational monitoring of the marine environment~\cite{SSHOPERATIONAL}. 
Furthermore, SSH characterization provides a plethora of data products that downstream tasks can use for many other applications~\cite{SSH3DCIRCULATION, 3DQGOC}.
Due to the irregular sampling delivered by satellite altimeter, state-of-the-art operational methods using optimal interpolation schemes~\cite{DUACS, MIOST} or model-driven data assimilation~\cite{DINEOF, DINEOF2, ANALOGDA, ANALOGDA2} fail to fully retrieve SSH dynamics at fine scales below 100-200km on a global or regional scale, so improving the space-time resolution of SSH fields has been a critical challenge in ocean science. 
Beyond some technological developments~\cite{SWOT}, recent studies support the critical role of ML-based schemes in overcoming the current limitations of the operational systems~\cite{4DVARNETSWOT, BFNQG, SSHInterpAttention} .  
The rest of this section gives an overview of the general problem definition for SSH interpolation, followed by a brief ontology for ML approaches to address the problem. 
We also give an overview of some experimental designs and datasets with a demonstration of metrics and plots generated by the \texttt{OceanBench} platform.

\subsection{Problem Definition}\label{sec:prob_definition}

We are dealing with satellite observations, so we are interested in the domain across the Earth's surface. 
Let us define the Earth's domain by some spatial coordinates, $\mathbf{x} = [\text{Longitude},\text{Latitude}]^\top \in\mathbb{R}^{D_s}$, and temporal coordinates, $t=[\text{Time}]\in\mathbb{R}^+$, where $D_s$ is the dimensionality of the coordinate vector.  
We can define some spatial (sub-)domain, $\Omega\subseteq\mathbb{R}^{D_s}$, and a temporal (sub-)domain, $\mathcal{T}\subseteq\mathbb{R}^+$. 
This domain could be the entire globe for 10 years or a small region within the North Atlantic for 1 year.
\begin{align}  \label{eq:spatiotemporal_coords}
    \text{Spatial Coordinates}: && \mathbf{x} &\in \Omega \subseteq \mathbb{R}^{D_s}\\ 
    \text{Temporal Coordinates}: && t &\in \mathcal{T} \subseteq \mathbb{R}^+.
\end{align}
In this case $D_s=2$ because we only have a two coordinates, however we can do some coordinate transformations like spherical to Cartesian. Likewise, we can do some coordinate transformation for the temporal coordinates like cyclic transformations or sinusoidal embeddings~\cite{ATTENTION}. We have two fields of interest from these spatiotemporal coordinates: the state and the observations.
\begin{align} \label{eq:state_obs}
    \text{State}: && \boldsymbol{u}(\mathbf{x},t) &: \Omega\times\mathcal{T}\rightarrow\mathbb{R}^{D_u} \\
    \text{Observations}: && \boldsymbol{y}_{obs}(\mathbf{x},t) &: \Omega\times\mathcal{T}\rightarrow\mathbb{R}^{D_{obs}}
\end{align}
The state domain, $u\in\mathcal{U}$, is a scalar or vector-valued field of size $D_u$ which is typically the quantity of interest and the observation domain, $y_{obs}\in\mathcal{Y}_{obs}$, is the observable quantity which is also a scalar or vector-valued field of size $D_{obs}$. Now, we make the assumption that we have an operator $\mathcal{H}$ that transforms the field from the state space, $\boldsymbol{u}$, to the observation space, $\boldsymbol{y}_{obs}$.
\begin{align} \label{eq:prob_definition}
    \boldsymbol{y}_{obs}(\mathbf{x},t) = \mathcal{H}\left(\boldsymbol{u}(\mathbf{x},t), t, \boldsymbol{\varepsilon}, \boldsymbol{\mu}\right) 
\end{align}
This equation is the continuous function defined over the entire spatiotemporal domain.  
The operator, $\mathcal{H}(\cdot)$, is flexible and problem dependent.
For example, in a some discretized setting there are 0's wherever there are no observations, and 1's wherever there are observations, and in other discretized settings it takes a weighted average of the neighboring pixels.
We also include a generic noise function, $\boldsymbol{\varepsilon}(\mathbf{x},t)$.
This could stem from a distribution, it could stationary noise operator, $\boldsymbol{\varepsilon}(\mathbf{x})$, or it could be constant in space but vary with Time, $\boldsymbol{\varepsilon}(t)$. 
We also include a control parameter, $\boldsymbol{\mu}$, representing any external factors or latent variables that could connect the state vector to the observation vector, e.g., sea surface temperature.
%
%
%
Our quantity of interest is SSH, $\eta$, a scalar-valued field defined everywhere on the domain. In our application, we assume that the SSH we observe from satellite altimeters, $\eta_{obs}$, is the same as the SSH state, except it could be missing for some coordinates due to incomplete coverage from the satellite. So our transformation is defined as follows:
\begin{align} \label{eq:ssh_field_continuous}
\boldsymbol{\eta}_{obs}(\mathbf{x},t) &= \mathcal{H}\left(\boldsymbol{\eta}(\mathbf{x},t), t, \boldsymbol{\varepsilon}, \boldsymbol{\mu}\right)
\end{align}
In practice, the satellite providers have a reasonable estimation of the amount of structured noise level we can expect from the satellite altimetry data; however, unresolved noise could still be present. 
Finally, we are interested in finding some model, $\mathcal{M}$, that maps the SSH we observe to the true SSH given by
\begin{align} \label{eq:interp_problem}
    \mathcal{M} &: \boldsymbol{\eta}_{obs}(\mathbf{x}, t, \boldsymbol{\mu}) \rightarrow \boldsymbol{\eta}(\mathbf{x},t),
\end{align}
which is essentially an inverse problem that maps the observations to the state.
One could think of it as trying to find the inverse operator, $\mathcal{M}=\mathcal{H}^{-1}$, but this could be some other arbitrary operator.  
\subsection{Machine Learning Model Ontology} \label{sec:ml_ontology_mini}

In general, we are interested in finding some parameterized operator, $\mathcal{M}_{\boldsymbol{\theta}}$, that maps the incomplete SSH field to the complete SSH field
\begin{align} \label{eq:ml_interp_problem}
    \mathcal{M}_{\boldsymbol{\theta}} &: \boldsymbol{\eta}_{obs}(\mathbf{x}, t, \boldsymbol{\mu}) \rightarrow \boldsymbol{\eta}(\mathbf{x},t),
\end{align}
whereby we learn the parameters from data.
The two main tasks we can define from this problem setup are 1) interpolation and 2) extrapolation.
We define \textit{interpolation} as the case when the boundaries of the inferred state domain lie within a predefined shape for the boundaries of the spatiotemporal observation domain. 
For example, the shape of the spatial domain could be a line, box, or sphere, and the shape of the temporal domain could be a positive real number line.
We define \textit{extrapolation} as the case where the boundaries of the inferred state domain are outside the boundaries of the spatiotemporal observation domain. 
In this case, the inferred state domain could be outside of either domain or both. 
A prevalent specific case of extrapolation is \textit{hindcasting} or \textit{forecasting}, where the inferred state domain lies within the spatial observation domain's boundaries but outside of the temporal observation domain's.
In the rest of this paper, we will look exclusively at the interpolation problem. 
However, we refer the reader to appendix~\ref{sec:other_tasks} for a more detailed look at other subtasks that can arise.

From a ML point of view, we can explore various ways to define the operator in equation~\eqref{eq:interp_problem}. 
We may distinguish three main categories: (i) coordinate-based methods that learn a parameterized continuous function to map the domain coordinates to the scalar values, (ii) the explicit mapping of the state from the observation, (iii) implicit methods defined as the solution of an optimization problem. 
The first category comprises of kriging approaches, which have been used operationally with historical success~\cite{KRIGINGREVIEW,DUACS}. Beyond such covariance-based approaches, recent contributions explore more complex trainable functional models~\cite{GPsBIGDATA}, basis functions~\cite{MIOST}, and neural networks~\cite{NERFSSSH}. 
The second category of schemes bypasses the physical modeling aspect and amortizes the prediction directly using state-of-the-art neural architectures such as UNets and ConvLSTMs~\cite{SSHInterpAttention, SSHInterpConvLSTM, SSHInterpUNet}. 
This category may straightforwardly benefit from available auxiliary observations~\citep{CDSOBSSST,CDSOBSSSTENS,CDSOBSOC} 
to state the interpolation problem as a super-resolution~\cite{SuperResSurvey} or image-to-image translation problem~\cite{IMAGE2IMAGETRANSLATION, IMAGE2IMAGETRANSLATION2}. 
The third category relates to inverse problem formulations and associated deep learning schemes, for example deep unfolding methods and plug-and-play priors~\cite{DEEPUNFOLDING}. 
Interestingly, recent contributions explore novel neural schemes which combine data assimilation formulations~\cite{DAGEOSCIENCE} and learned optimizer strategies~\cite{4DVARNETSWOT,4DVARNETSST}.
We provide a more detailed ontology of methods used for interpolation problems in appendix~\ref{sec:ml_ontology}. 
We consider at least one baseline approach from each category for each data challenge described in section~\ref{sec:data_challenges}. 
While all these methods have pros and cons, we expect the OceanBench platform to showcase to new experimental evidence and understanding regarding their applicability to SSH interpolation problems.

\subsection{Experimental Design} \label{sec:experimental_design}

\begin{table}[ht]
\caption{This table gives a brief overview of the datasets provided to complete the data challenges listed in~\ref{sec:data_challenges} and~\ref{sec:data_challenges_extended}. Note that the OSSE datasets are all gridded products whereas the OSE NADIR is an alongtrack product. See figure~\ref{fig:oceanbench_maps} for an example of the OSSE NEMO Simulations for SSH and SST and pseudo-observations for NADIR \& SWOT.}
\label{tb:datasets}
\centering
\begin{tabular}{lcccc}
 \toprule
 & OSSE & OSSE NADIR + SWOT & OSSE SST & OSE NADIR  \\ \midrule
 Data Type & Simulations & 
Pseudo-Observations & 
 Simulations & Observations \\
Source     & 
NEMO~\citep{NEMOAJAYI2020} & 
NEMO~\citep{NEMOAJAYI2020} &
NEMO~\citep{NEMOAJAYI2020}
& Altimetry~\citep{MDSALONGTRACK} \\
Region & 
GulfStream & GulfStream & GulfStream & GulfStream \\
Domain Size &
$10\times 10^\circ$ &
$10\times 10^\circ$ &
$10\times 10^\circ$ &
$10\times 10^\circ$
\\
Longitude Extent &
$[-65^\circ, -55^\circ]$ & 
$[-65^\circ, -55^\circ]$ &
$[-65^\circ, -55^\circ]$ &
$[-65^\circ, -55^\circ]$ \\
Latitude Extent &
$[33^\circ, 43^\circ]$ &
$[33^\circ, 43^\circ]$ &
$[33^\circ, 43^\circ]$ &
$[33^\circ, 43^\circ]$ \\
Resolution &
$0.05^\circ\times 0.05^\circ$ &
$0.05^\circ\times 0.05^\circ$ &
$0.05^\circ\times 0.05^\circ$ &
$7$ km \\
Grid Size &
$200\times 200$ & $200\times 200$ & $200\times 200$ & N/A \\
Num Datapoints &
$\sim$14.6M & $\sim$14.6M & $\sim$14.6M & $\sim$1.6M \\
Period Start & 2012-10-01 & 2012-10-01 & 2012-10-01 & 2016-12-01 \\
Period End & 2013-09-30 & 2013-09-30 & 2013-09-30 & 2018-01-31 \\
Frequency  & Daily & Daily & Daily & 1 Hz \\
\bottomrule
\end{tabular}
\end{table}

The availability of multi-year simulation and observation datasets naturally advocates for the design of synthetic (or twin) experiments, referred to as observing system simulation experiments (OSSE), and of real-world experiments, referred to as observing system experiments (OSE).
We outline these two experimental setups below.

\textbf{Observing System Simulation Experiments (OSSE)}. A staple and groundtruthed experimental setup uses a reference simulation dataset to simulate the conditions we can expect from actual satellite observations. 
This setup allows researchers and operational centers to create a fully-fledged pipeline that mirrors the real-world experimental setting.
An ocean model simulation is deployed over a specified spatial domain and period, and a satellite observation simulator is deployed to simulate satellite observations over the same domain and period. 
This OSSE setup has primarily been considered for performance evaluation, as one can assess a reconstruction performance over the entire space-time domain. It also provides the basis for the implementation of classic supervised learning strategies~\cite{SSHInterpUNet,SSHInterpConvLSTM,SSHInterpAttention}.
The domain expert can vary the experimental conditions depending on the research question. 
For example, one could specify a region based on the expected dynamical regime~\cite{DCOSSEGULFSSH} or add a certain noise level to the observation tracks based on the satellite specifications.
The biggest downside to OSSE experiments is that we train models exclusively with ocean simulations which could produce models that fail to generalize to the actual ocean state. 
Furthermore, the simulations are often quite expensive, which prevents the community from having high spatial resolution over very long periods, which would be essential to capture as many dynamical regimes as possible.

\begin{figure}[t!]
\small
\begin{center}
\setlength{\tabcolsep}{1pt}
\begin{tabular}{ccc}
NADIR Altimetry Tracks & 
SWOT Altimetry Tracks &
Sea Surface Temperature \\
\includegraphics[width=42.5mm, height=30mm]{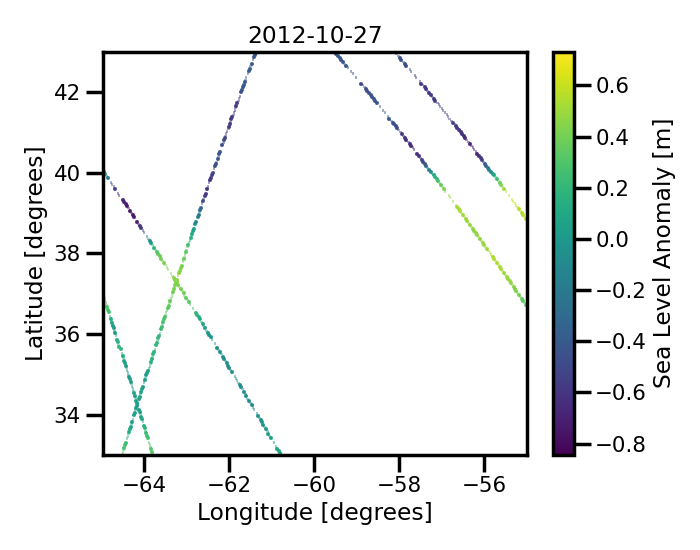} 
&
\includegraphics[width=42.5mm, height=30mm]{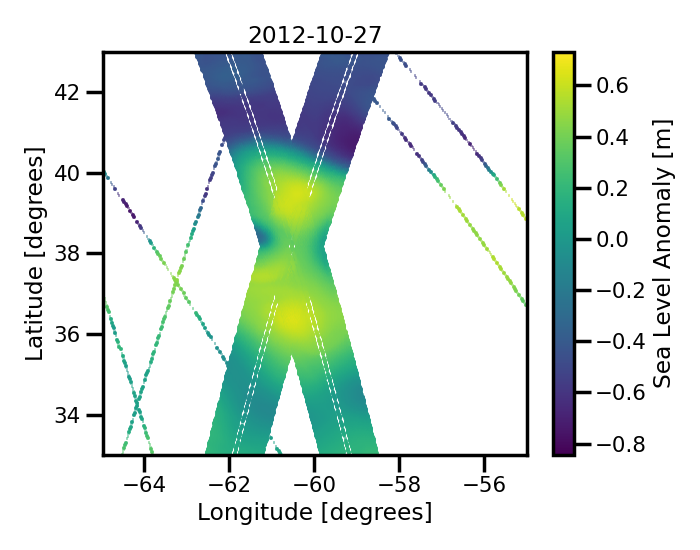} &
\includegraphics[width=4.25cm,height=3cm]{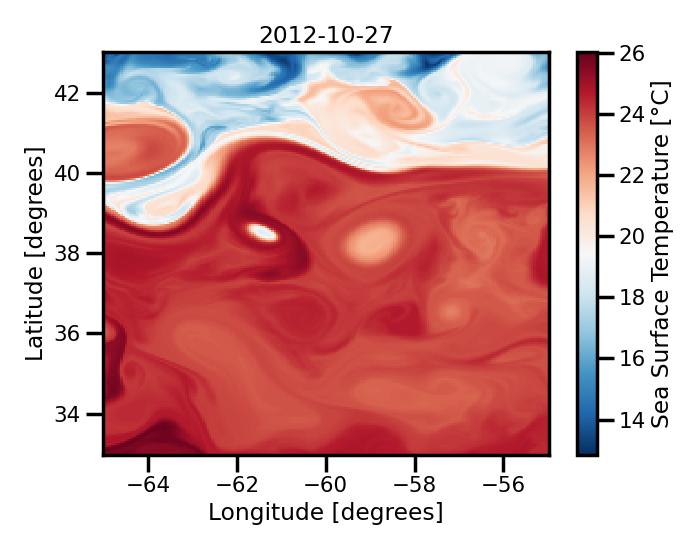}
\end{tabular}
\begin{tabular}{cccc}
\hspace{3mm} NEMO Simulation & 
\hspace{3mm} MIOST & 
\hspace{3mm} BFNQG & 
4DVarNet \\
\vspace{-2mm}
\includegraphics[trim={0 0 42mm 0},clip, width=3.20cm,height=3cm]{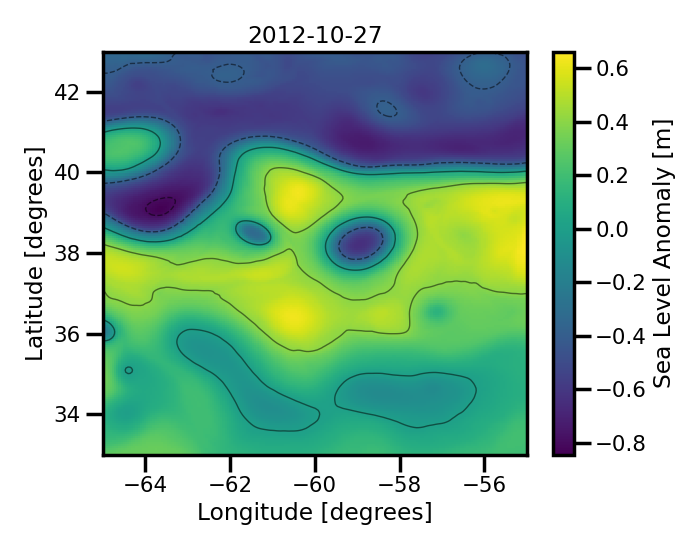} &
\includegraphics[trim={0 0 42mm 0},clip, width=3.2cm,height=3cm]{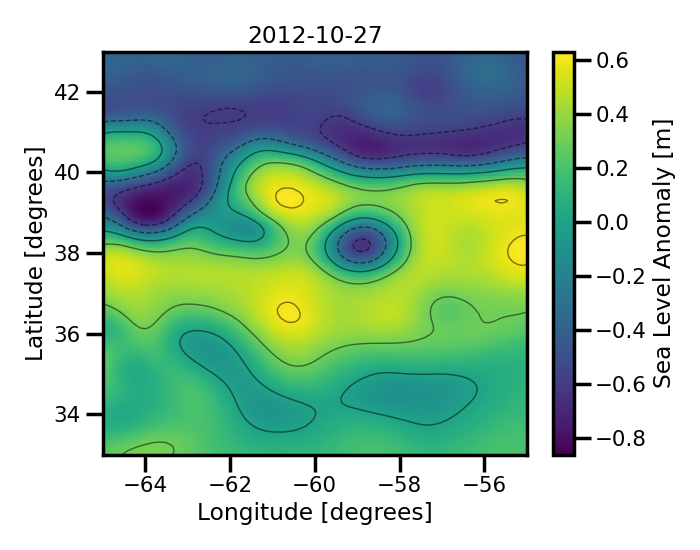} &
\includegraphics[trim={0 0 42mm 0},clip, width=3.2cm,height=3cm]{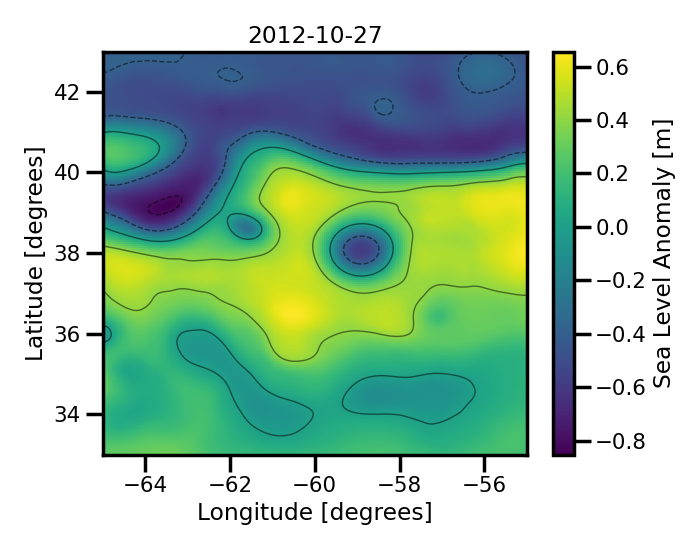} &
\includegraphics[width=4.0cm,height=3cm]{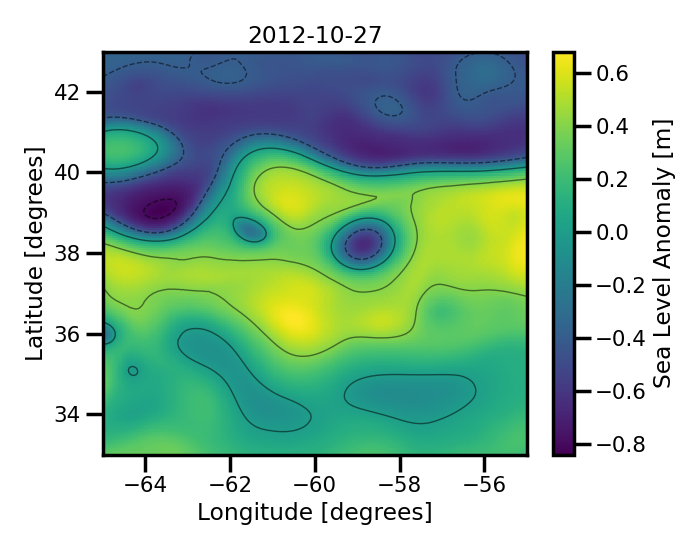} \\
\vspace{-2mm}
\includegraphics[trim={0 0 42mm 0},clip, width=3.20cm,height=3cm]{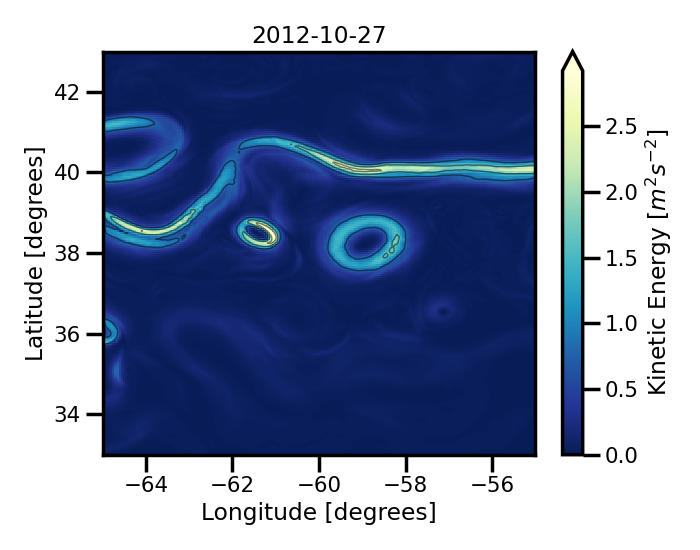} &
\includegraphics[trim={0 0 42mm 0},clip, width=3.2cm,height=3cm]{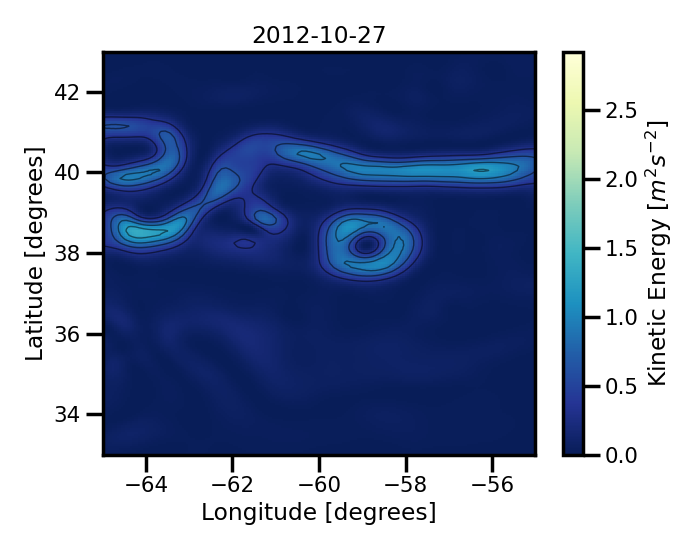} &
\includegraphics[trim={0 0 42mm 0},clip, width=3.2cm,height=3cm]{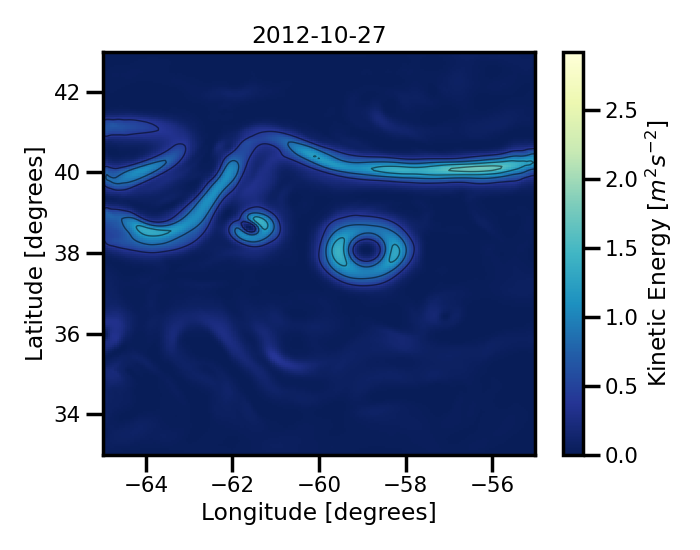} &
\includegraphics[width=4.0cm,height=3cm]{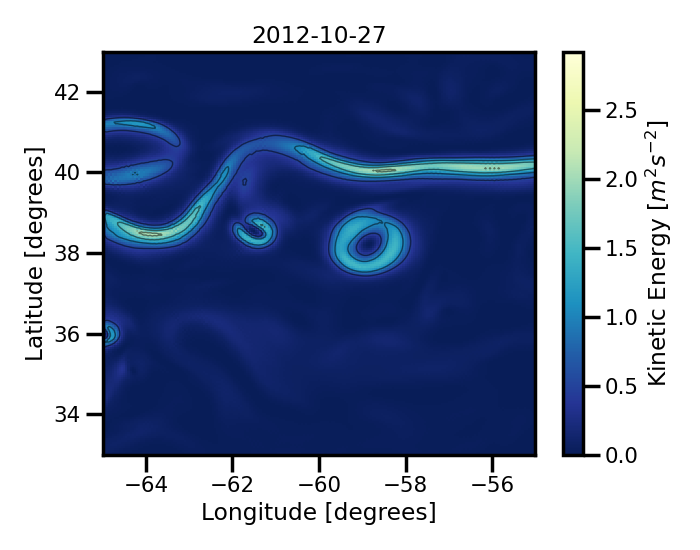}  \\
\vspace{-2mm}
\includegraphics[trim={0 0 42mm 0},clip, width=3.20cm,height=3cm]{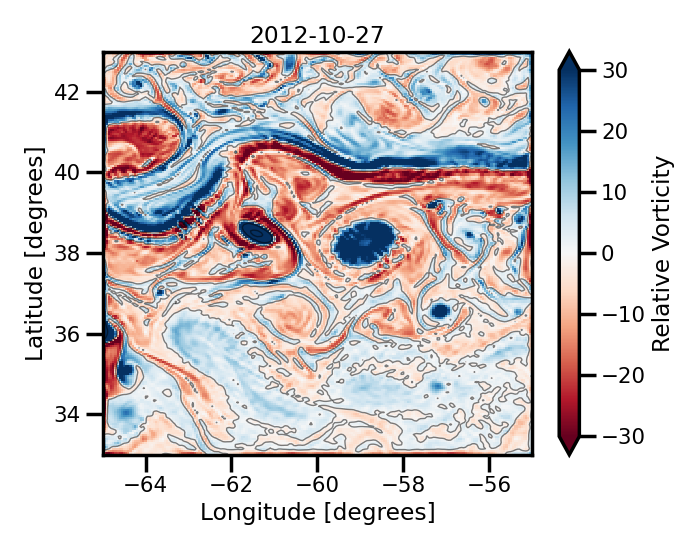} &
\includegraphics[trim={0 0 42mm 0},clip, width=3.2cm,height=3cm]{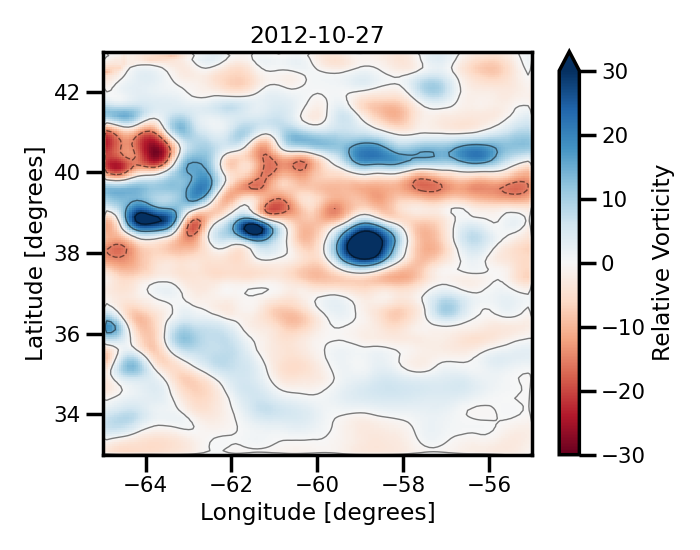} &
\includegraphics[trim={0 0 42mm 0},clip, width=3.2cm,height=3cm]{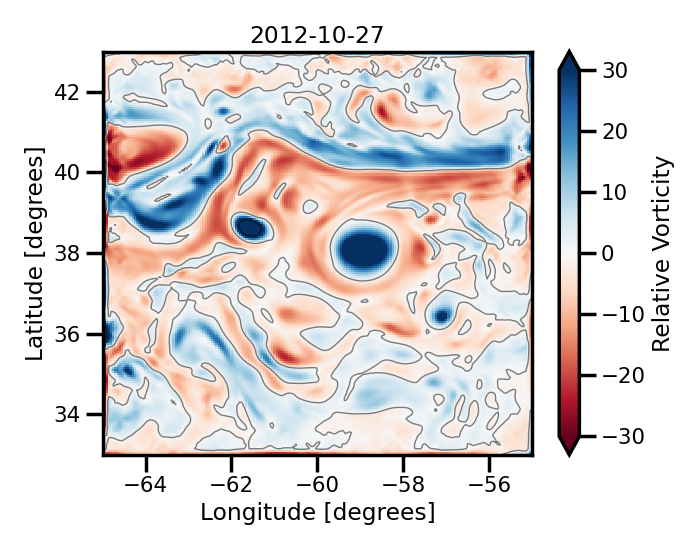} &
\includegraphics[width=4.0cm,height=3cm]{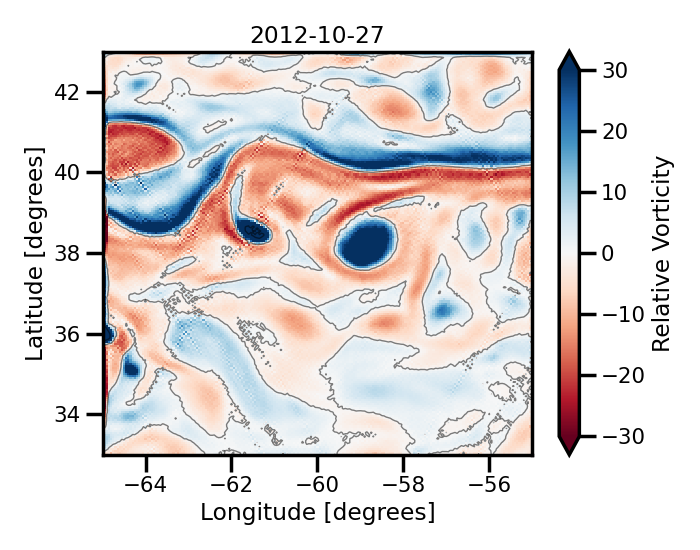}  \\
\includegraphics[trim={0 0 38mm 0},clip, width=3.20cm,height=3cm]{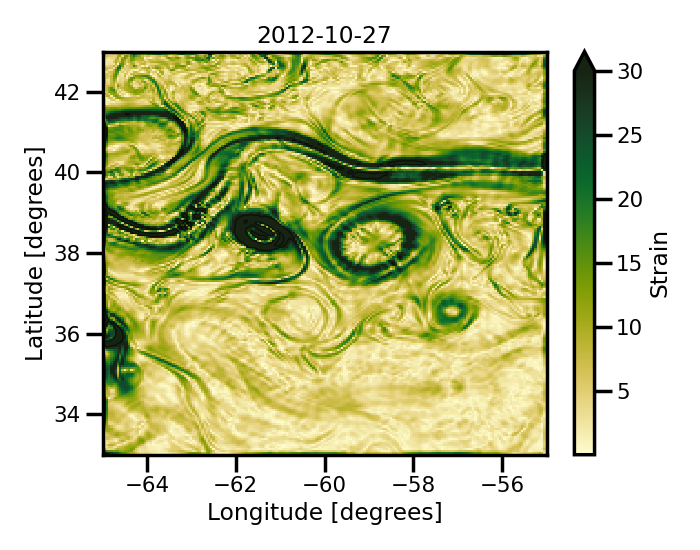} &
\includegraphics[trim={0 0 38mm 0},clip, width=3.2cm,height=3cm]{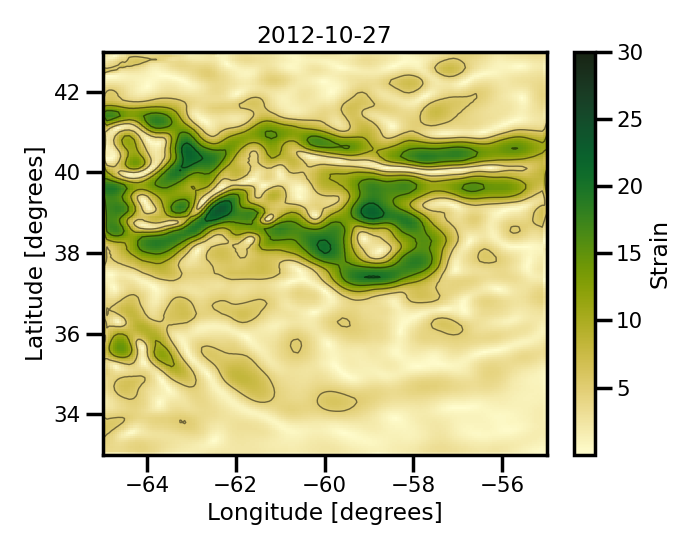} &
\includegraphics[trim={0 0 38mm 0},clip, width=3.2cm,height=3cm]{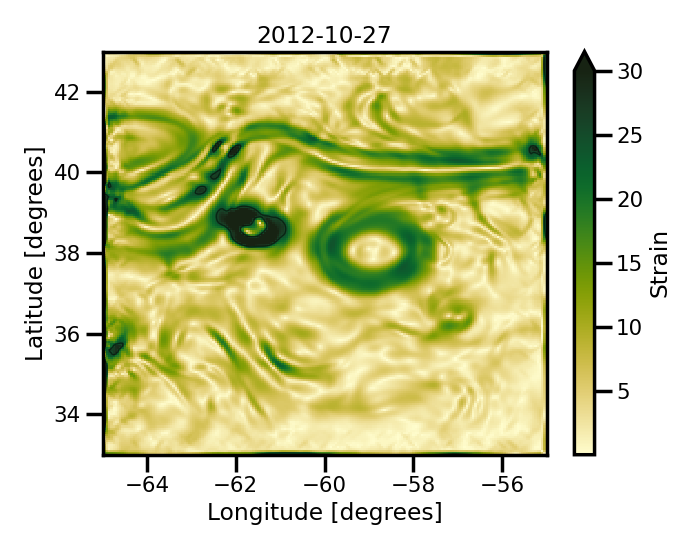} &
\includegraphics[width=4.0cm,height=3cm]{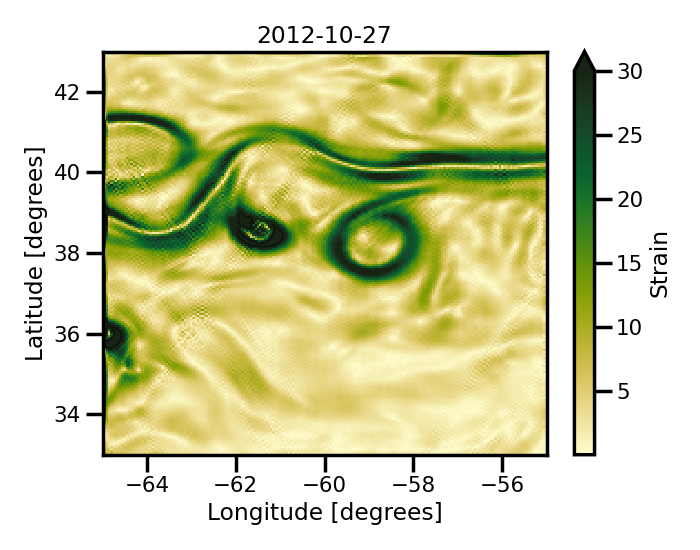}  \\
(a) & (b) & (c) & (d)
\end{tabular}
\vspace{-3mm}
\caption{
A snapshot at $27^{th}$ October, 2012 of the sea level anomaly (SLA) from the NEMO simulation for the OSSE experiment outlined in section~\ref{sec:experimental_design}. 
The top row showcases the aggregated NADIR altimetry tracks and the aggregated SWOT altimetry tracks (12 hours before and 12 hours after) as well as the SST from the NEMO simulation.
Each subsequent row showcases the following physical variables found in appendix~\ref{sec:physical_variables}: (a) Sea Level Anomaly, (b) Kinetic Energy, (c) Relative Vorticity, and (d) Strain. 
Each column in the subsequent rows showcase the following reconstructed field from the NEMO simulation found in columrn (a): (b) MIOST~\cite{MIOST}, (c) BFN-QG~\cite{BFNQG}, and (d) 4DVarNet~\cite{4DVARNETSWOT}.}
\vspace{-5mm}
\label{fig:oceanbench_maps}
\end{center}
\end{figure}


\textbf{Observing System Experiments (OSE)}. As more observations have become available over the past few decades, we can also design experiments using real data. 
This involves aggregating as many observations from real ocean altimetry satellites as possible with some specific independent subset left out for evaluation purposes.
A major downside to OSE experiments is that the sparsity and spatial coverage of the observations narrow the possible scope of performance metrics and make it very challenging to learn directly from observation datasets. 
The current standard altimetry data are high resolution but cover a tiny area. 
As such, it can only inform fine-scale SSH patterns in the along-track satellite direction and cannot explicitly reveal two-dimensional patterns. 
Despite these drawbacks, it provides a quantitative evaluation of the generalizability of the ML methods concerning the true ocean state.

\begin{figure}[t!]
\small
\begin{center}
\setlength{\tabcolsep}{2pt}
\begin{tabular}{ccc}
\includegraphics[width=3.75cm,height=3.25cm]{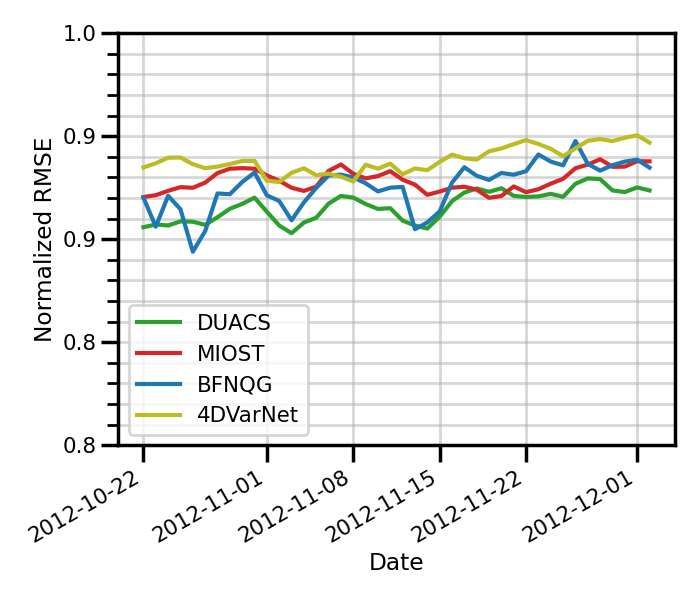} &
\includegraphics[width=4.25cm,height=3.5cm]{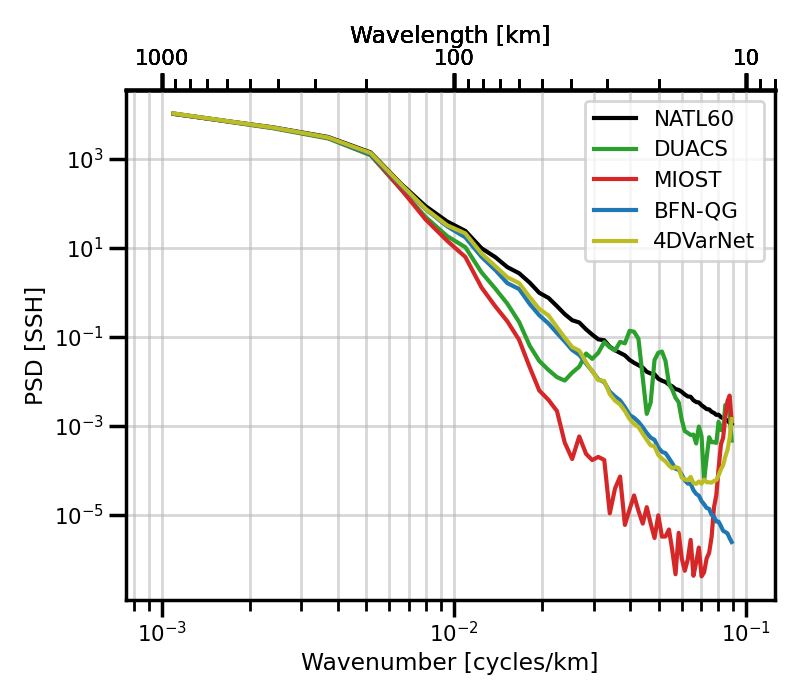} &
\includegraphics[width=4.25cm,height=3.5cm]{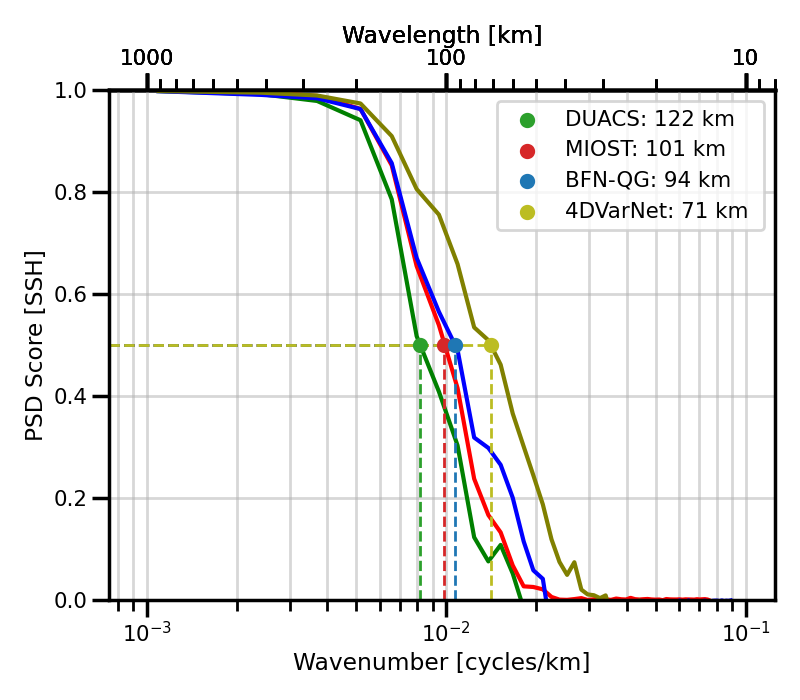} 
\\
(a) Normalized RMSE &
(b) Isotropic Power Spectrum &
(c) Isotropic Power Spectrum Score
\end{tabular}
\begin{tabular}{cccc}
\includegraphics[trim={0 0 0mm 0},clip, width=4.20cm,height=3cm]{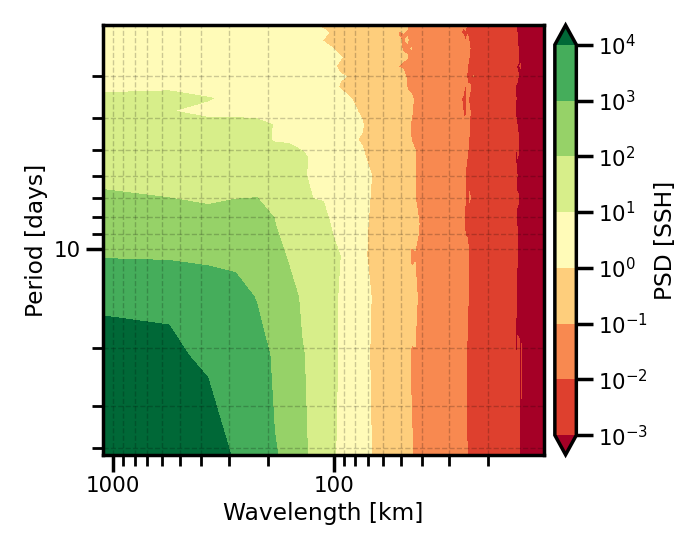}  &
\includegraphics[trim={20mm 0 34mm 0},clip, width=2.9cm,height=3cm]{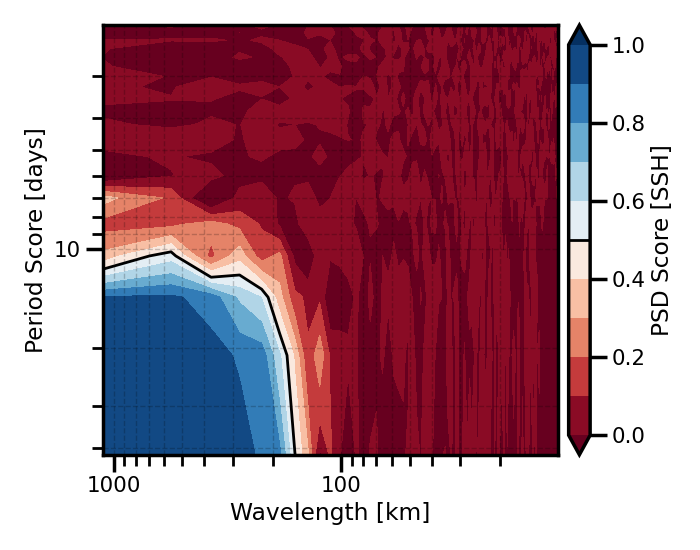} &
\includegraphics[trim={20mm 0 34mm 0},clip, width=2.9cm,height=3cm]{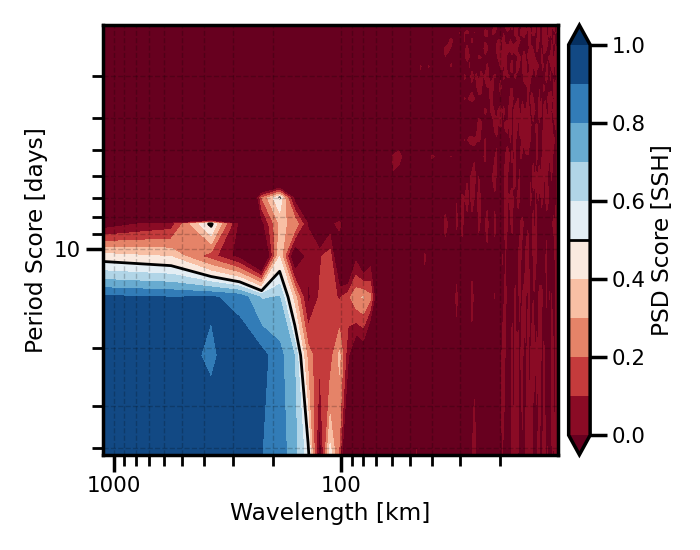} &
\includegraphics[trim={20mm 0 0 0},clip, width=3.5cm,height=3cm]{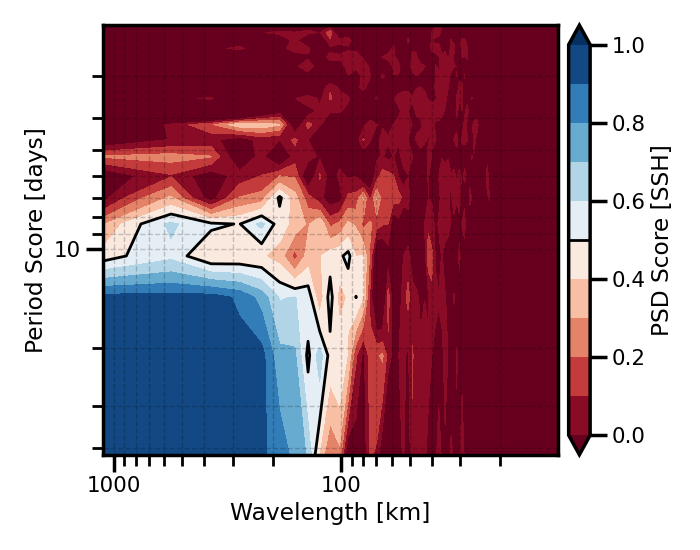} \\
(d) NEMO Simulation &
(e) MIOST &
(f) BFN-QG &
(g) 4DVarNet
\end{tabular}
\caption{This figure showcases some statistics for evaluation of the SSH field reconstructions for the OSSE NADIR experiment outlined in section~\ref{sec:interp_challenge}. Subfigure (a) showcases the normalized root mean squared error (nRMSE), (b) showcases the isotropic power spectrum decomposition (PSD), (c) showcases isotropic PSD scores.
The bottom row showcases the space-time PSD for the NEMO simulation (subfigure (d)) and the PSD scores for three reconstruction models: (e) the MIOST model~\cite{MIOST}, (f) the BFN-QG model~\cite{BFNQG}, and (g) the 4DVarNet model~\cite{4DVARNETSWOT}.
}
\label{fig:oceanbench_psd}
\end{center}
\end{figure}

\subsection{Data Challenges} \label{sec:data_challenges}

We rely on existing OSSE and OSE experiments for SSH interpolation designed by domain experts~\cite{DCOSEGULFSSH,DCOSSEGULFSSH} and recast them into \texttt{OceanBench} framework to deliver a ML-ready benchmarking suites. 
The selected data challenges  for this first edition address SSH interpolation for a 1000km$\times$1000km Gulfstream region. We briefly outline them below.

\textbf{Experiment I (\textit{OSSE NADIR})} addresses SSH interpolation using NADIR altimetry tracks which are very fine, thin ocean satellite observations (see Figure~\ref{fig:oceanbench_maps}). It relies on an OSSE using high-resolution ($1/60^\circ$ resolution) ocean simulations generated by the NEMO model over one year with a whole field every day. 

\textbf{Experiment II (\textit{OSSE SWOT})} addresses SSH interpolation using jointly NADIR and SWOT altimetry data where we complement the \textbf{OSSE NADIR} configuration with simulated SWOT observations.
SWOT is a new satellite altimetry mission with a much higher spatial coverage but a much lower temporal resolution as illustrated in Figure~\ref{fig:oceanbench_maps}.
The higher spatial resolution allows us to see structures at a smaller resolution but at the cost of a massive influx of observations (over $\times$100).

\textbf{Experiment III (\textit{OSSE SST})} addresses SSH interpolation using altimetry and SST satellite data jointly. We complement the \textbf{OSSE SWOT} challenge with simulated SST observations. 
Satellite-derived SST observations are more abundantly available in natural operational settings than SSH at a finer resolution, and structures have visible similarities~\cite{SWOT,BFNQG}.
So this challenge allows for methods to take advantage of multi-modal learning~\cite{4DVARNETSST,SSHInterpAttention}.

\textbf{Experiment IV (\textit{OSE NADIR})} addresses SSH interpolation for real NADIR altimetry data. 
In contrast to the three OSSE data challenges, it only looks at actual observations aggregated from the currently available ocean altimetry data from actual satellites. 
It involves a similar space-time sampling as Experiment (\textbf{OSSE NADIR}) to evaluate the generalization of ML methods trained in Experiment I to real altimetry data. 
The training problem's complexity increases significantly due to the reference dataset's sparsity compared with the \textbf{OSSE NADIR} dataset. 
One may also explore transfer learning or fine-tuning strategies from the available OSSE dataset.

\subsection{\texttt{OceanBench} Pipelines}

\begin{table}[h]
\caption{This table highlights some of the results for the \textbf{OSSE NADIR} experiment outlined in section~\ref{sec:data_challenges} and appendix~\ref{sec:data_challenges_extended}.
This table highlights the performance statistically in the real and spectral space; the normalized RMSE score for the real space and the minimum spatial and temporal scales resolved in the spectral domain. 
For more information about the class of models displayed and class of metrics, see appendix~\ref{sec:ml_ontology} and appendix~\ref{sec:metrics} respectively. We only showcase the model performance on the alongtrack NADIR data available. For the extended table for each of the challenges, see Table~\ref{tb:exp-results-mega}.}
\label{tb:oceanbench_results}
\centering
\begin{tabular}{lllcccc}
 \toprule
Experiment &  Algorithm &   Algorithm Class &  nRMSE Score & $\lambda_{\mathbf{x}}$ [km]  & $\lambda_{t}$ [days]      \\ \midrule
\multicolumn{1}{l}{OSSE NADIR}     &  OI~\cite{DUACS} &  Coordinate-Based & 0.92 $\pm$ 0.01 & 175 & 10.8 \\
\multicolumn{1}{l}{OSSE NADIR}     &  MIOST~\cite{MIOST} &  Coordinate-Based  & 0.93 $\pm$ 0.01 & 157 & 10.1 \\
\multicolumn{1}{l}{OSSE NADIR}     &  BFNQG~\cite{BFNQG} &  Hybrid Model   & 0.93 $\pm$ 0.01 & 139 & 10.6 \\
OSSE NADIR &  4DVarNet~\cite{4DVARNETSWOT} &  Bi-Level Opt.  & 0.95 $\pm$ 0.01 & 117 & 7.7 \\
\bottomrule
\end{tabular}
\end{table}

For the four data challenges presented in the previous section, we used \texttt{OceanBench} pipelines to deliver a ML-ready benchmarking framework.
We used the \texttt{hydra} and the geoprocessing tools outlined in section~\ref{sec:code_structure} with specialized routines for regridding the ocean satellite data to a uniformly gridded product and vice versa when necessary. 
Appendix~\ref{sec:hydra_recipes} showcases an example of the hydra integration for the preprocessing pipeline. 
A key feature is the creation of a custom patcher for the appropriate geophysical variables using our \texttt{XRPatcher} tool, which is later integrated into custom datasets and dataloaders for the appropriate model architecture, e.g., coordinate-based or grid-based. 
We provide an example snippet of how this can be done easily in section~\ref{sec:xrpatcher}.
\texttt{OceanBench} also features some tools specific to the analysis of SSH. 
For example, physically-interpretable variables like geostrophic currents and relative vorticity, which can be derived from first-order and second-order derivatives of the SSH, are essential for assessing the quality of the reconstructions generated by the models. 
Figure~\ref{fig:oceanbench_maps} showcases some fields of the most common physical variables used in the oceanography literature for the SSH-based analysis of sea surface dynamics. For more details regarding the nature of the physical variables, see appendix~\ref{sec:physical_variables}.

Regarding the evaluation framework, we include domain-relevant performance metrics beyond the standard ML loss and accuracy functions. They account for the sampling patterns of the evaluation data. Spectral analytics are widely used in geoscience~\cite{BFNQG}, and here, we consider spectral scores computed as the minimum spatial and temporal scales resolved by the reconstruction methods proposed in~\cite{BFNQG}.
For example, figure~\ref{fig:oceanbench_psd} showcases how \texttt{OceanBench} generated the isotropic power spectrum and score and the space-time power spectrum decomposition and score.
Table~\ref{tb:oceanbench_results} outlines some standard and domain-specific scores for the experiments outlined in section~\ref{sec:experimental_design}.
We give a more detailed description of the rationale and construction of the power-spectrum-specific metrics in appendix~\ref{sec:metrics}. In terms of baselines, we report for each data challenge the performance of at least one approach for each of the category outlined in Section \ref{sec:ml_ontology_mini}.

\section{Conclusions} \label{sec:conclusions}

The ocean community faces technological and algorithmic challenges to make the most of available observation and simulation datasets. 
In this context, recent studies evidence the critical role of ML schemes in reaching breakthroughs in our ability to monitor ocean dynamics for various space-time scales and processes. 
Nevertheless, domain-specific preprocessing steps and evaluation procedures slow down the uptake of ML toward real-world applications. 
Our application of choice was SSH mapping which facilities the production of many crucial derived products that are used in many downstream tasks like subsequent modeling~\citep{ML4OCN}, ocean health monitoring~\citep{ML4NATURECONSERVATION,OCNHEALTH,OCEANHEALTH2} and maritime risk assessment~\citep{SSHOPERATIONAL}.

Through \texttt{OceanBench} framework, we embed domain-level requirements into the MLOPs considerations by building a flexible framework that adds this into the hyperparameter considerations for ML models. 
We proposed four challenges towards a ML-ready benchmarking suite for ocean observation challenges. 
We outlined the inner workings \texttt{OceanBench} and demonstrated its usefulness by recreating some preprocessing and analysis pipelines from a few data challenges involving SSH interpolation.
We firmly believe that the \texttt{OceanBench} platform is a crucial step to lowering the barrier of entry for new ML researchers interested in applying and developing their methods to relevant problems in the ocean sciences.

\newpage
\begin{ack}
This work was supported by the French National Research Agency (ANR), through projects number ANR-17- CE01-0009-01, ANR-19-CE46-0011 and ANR-19-CHIA-0016); by the French National Space Agency (CNES) through the SWOT Science Team program (projects MIDAS and DIEGO) and the OSTST program (project DUACS-HR); by the French National Centre for Scientific Research (CNRS) through the LEFE-MANU program (project IA-OAC). This project also received funding from the European Union’s Horizon Europe research and innovation programme under the grant No 101093293 (EDITO-Model Lab project). This project benefited from HPC and GPU computing resources from GENCI-IDRIS (Grant 2021-101030).
\end{ack}
\section*{Checklist}


\begin{enumerate}

\item For all authors...
\begin{enumerate}
  \item Do the main claims made in the abstract and introduction accurately reflect the paper's contributions and scope?
    \answerYes{All the contributions listed in the abstract are elaborated in sections~\ref{sec:code_structure},~\ref{sec:data_challenges} and~\ref{sec:conclusions}}
  \item Did you describe the limitations of your work?
    \answerYes{See the last paragraph of section 5 and the appendix as well.}
  \item Did you discuss any potential negative societal impacts of your work?
    \answerYes{We do not believe that our work has any potential negative societal impacts directly as we do not deal with any confidential or private data. However, we do outline in the appendix how there may be some adverse effects related to downstream uses which could have some negative societal impacts.}
  \item Have you read the ethics review guidelines and ensured that your paper conforms to them?
    \answerYes{We do not include any confidential or private data. We only include numerical values which stem from general physical systems or machine learning models. We do not believe they hold any ethical issues. However, we do acknowledge that there would be environmental damage should users go forward and explore methods which obscenely high computing hours. This discussion outlined in the appendix.}
\end{enumerate}

\item If you are including theoretical results...
\begin{enumerate}
  \item Did you state the full set of assumptions of all theoretical results?
    \answerNA{We do not include any theoretical results.}
	\item Did you include complete proofs of all theoretical results?
    \answerNA{We do not include any theoretical results.}
\end{enumerate}

\item If you ran experiments (e.g. for benchmarks)...
\begin{enumerate}
  \item Did you include the code, data, and instructions needed to reproduce the main experimental results (either in the supplemental material or as a URL)?
    \answerYes{We include the parameters used to reproduce the dataset preprocessing and evaluation procedure in Appendix \ref{sec:data_challenges_extended} and instructions are given to download the data via~\href{https://github.com/quentinf00/oceanbench-data-registry}{https://github.com/quentinf00/oceanbench-data-registry} and rerun the evaluation procedure in our code repository which is available at~\href{https://github.com/jejjohnson/oceanbench}{https://github.com/jejjohnson/oceanbench}.}
  \item Did you specify all the training details (e.g., data splits, hyperparameters, how they were chosen)?
    \answerYes{We showcase all preprocessing steps necessary to reproduce the experimental configurations in Appendix~\ref{sec:data_challenges_extended} and the configuration files are available in our code repository at~\href{https://github.com/jejjohnson/oceanbench}{https://github.com/jejjohnson/oceanbench}. }
	\item Did you report error bars (e.g., with respect to the random seed after running experiments multiple times)?
    \answerNA{This is not applicable for this instantiation because we do not include any randomness within the experiment procedure nor the results.}
	\item Did you include the total amount of compute and the type of resources used (e.g., type of GPUs, internal cluster, or cloud provider)?
    \answerYes{We do not do any model training and leave it up the user for their local or cloud machine. However, we do provide the cloud provider for the data found the the data registry which can be found at~\href{https://github.com/quentinf00/oceanbench-data-registry}{https://github.com/quentinf00/oceanbench-data-registry}}
\end{enumerate}

\item If you are using existing assets (e.g., code, data, models) or curating/releasing new assets...
\begin{enumerate}
  \item If your work uses existing assets, did you cite the creators?
    \answerYes{We adopted the implementation of the preprocessing procedures and evaluation steps with some modifications. We give proper citation and credit to the authors as well as all other existing software packages included in this work.}
  \item Did you mention the license of the assets?
    \answerYes{The appropriate license notices are included in the source code files.}
  \item Did you include any new assets either in the supplemental material or as a URL?
    \answerYes{All the processing and evaluation scripts are included in the GitHub repository.}
  \item Did you discuss whether and how consent was obtained from people whose data you're using/curating?
    \answerYes{We only include data that is already publicly available. We also discussed with the original generators of the datasets and keep the appropriate licenses.}
  \item Did you discuss whether the data you are using/curating contains personally identifiable information or offensive content?
    \answerNA{We do not include any personal information or offensive content in our datasets.}
\end{enumerate}

\item If you used crowdsourcing or conducted research with human subjects...
\begin{enumerate}
  \item Did you include the full text of instructions given to participants and screenshots, if applicable?
    \answerNA{We do not use crowdsourcing and we do not conduct research with human subjects.}
  \item Did you describe any potential participant risks, with links to Institutional Review Board (IRB) approvals, if applicable?
    \answerNA{See the previous point.}
  \item Did you include the estimated hourly wage paid to participants and the total amount spent on participant compensation?
    \answerNA{See the previous point.}
\end{enumerate}

\end{enumerate}

\bibliographystyle{plain}
\bibliography{
content/biblio, 
content/bibliographies/software, 
content/bibliographies/machine_learning, 
content/bibliographies/sea_surface_height,
content/bibliographies/ocean,
content/bibliographies/data,
content/bibliographies/applications
}







\newpage
\appendix
\section*{\textsc{OceanBench}: The Sea Surface Height Edition - Supplementary Material}
\section{Data Challenges} \label{sec:data_challenges_extended}

In this section, we highlight some details that were omitted in section~\ref{sec:data_challenges}.
This includes details about the simulation type, the data structures, and the training/evaluation periods.

\subsection{OSSE NADIR} \label{sec:osse_nadir}

The reference simulation is the \textit{NATL60} simulation based on the NEMO model~\cite{NEMOAJAYI2020}. 
This particular simulation was run over an entire year without any tidal forcing.
The simulation provides the outputs of SSH, SST, sea surface salinity (SSS) and the u,v velocities every 1 hour.
For the purposes of this data challenge, the spatial domain is over the Gulfstream with a spatial domain of $[-65^\circ, -55^\circ]$ longitude and $[33^\circ, 43^\circ]$ latitude.
The resolution of the original simulation is 1/60$^\circ$ resolution with hourly snapshots, and we consider a daily downsampled trajectory at 1/20$^\circ$ for the data challenge which results in a 365x200x200 spatio-temporal grid.
This simulation resolves finescale dynamical processes ($\sim$15km) which makes it a good test bed for creating an OSSE environment for mapping.
The SSH observations include simulations of ocean satellite NADIR tracks.
In particular, they are simulations of Topex-Poseidon, Jason 1, Geosat Follow-On, and Envisat.
There is no observation error considered within the challenge.
We use a the entire period from 2012-10-10 until 2013-09-30.
A training period is only from 2013-01-02 to 2013-09-30 where the users can use the reference simulation as well as all available simulated observations.
The evaluation period is from 2012-10-22 to 2012-12-02 (i.e. 41 days) which is considered decorrelated from the training period. 
During the evaluation period, the user cannot use the reference NATL60 simulation but they can use all available simulated observations. There is also a spin-up period allowance from 2012-10-01 where the user can also use all available simulated observations.

\subsection{OSSE SWOT \& OSSE SST} \label{sec:osse_swot_sst}

For the OSSE SWOT and OSSE SST experiments, the reference simulation, domain, and evaluation period is the same as the OSSE NADIR experiment.
However, the OSSE SWOT includes simulated observations of the novel KaRIN sensor recently deployed during the SWOT mission, the pseudo-observations were generated using the SWOT simulator~\cite{SWOT}. 
This OSSE SST experiment allows the users to utilize the full fields of SST as inputs to help reconstruct the SSH field in conjunction with the NADIR and SWOT SSH observation.
Because the SST comes from the same NATL60 simulation, the geometry characteristics SST and SSH are exactly the same.

\subsection{OSE NADIR} \label{sec:ose_nadir}

The OSE NADIR experiment only uses real observations aggregated from different altimeters. These SSH observations include observations from the SARAL/Altika, Jason 2, Jason 3, Sentinel 3A, Haiyang-2A and Cryosat-2 altimeters. The Cryosat-2 altimeter is used as the independent evaluation track used to assess the performance of the reconstructed SSH field.

\subsection{Results}

\begin{figure}[ht!]
\small
\begin{center}
\setlength{\tabcolsep}{1pt}
\begin{tabular}{cccc}
\hspace{3mm} Task OSSE & 
\hspace{3mm} Task OSSE & 
\hspace{2mm} Task OSSE & 
Task OSE \\
\hspace{3mm}  Nadir & 
\hspace{3mm} Nadir + SWOT & 
\hspace{2mm} Nadir + SST & 
Nadir \\
\includegraphics[trim={0 13mm 22mm 0},clip, width=3.60cm,height=3.2cm]{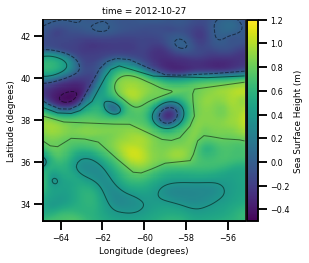} &
\includegraphics[trim={13mm 13mm 22mm 0},clip, width=3.2cm,height=3.2cm]{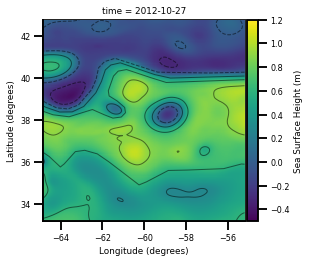} &
\includegraphics[trim={13mm 13mm 22mm 0},clip, width=3.2cm,height=3.2cm]{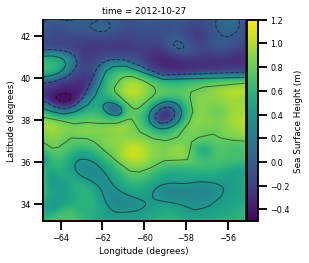} &
\includegraphics[trim={13mm 13mm 0 0},clip,width=4.0cm,height=3.2cm]{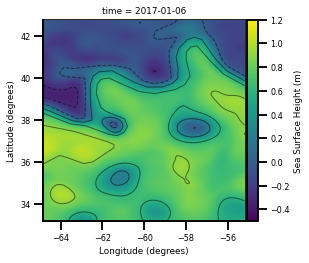} \\
\includegraphics[trim={0 13mm 22mm 5mm}, clip, width=3.60cm,height=3cm]{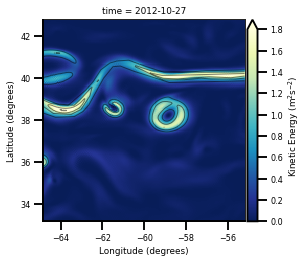} &
\includegraphics[trim={13mm 13mm 22mm 5mm},clip, width=3.2cm,height=3cm]{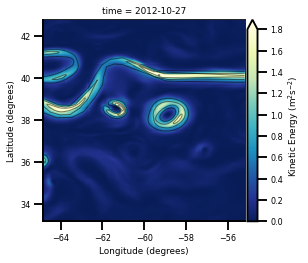} &
\includegraphics[trim={13mm 13mm 22mm 5mm},clip, width=3.2cm,height=3cm]{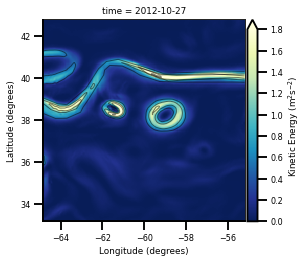} &
\includegraphics[trim={13mm 13mm 0 5mm},clip,width=4cm,height=3cm]{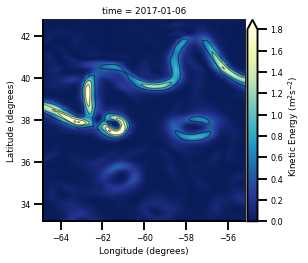} \\
\includegraphics[trim={0 13mm 21.2mm 5mm},clip, width=3.60cm,height=3cm]{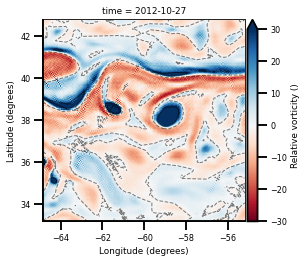} &
\includegraphics[trim={13mm 13mm 21.2mm 5mm},clip, width=3.2cm,height=3cm]{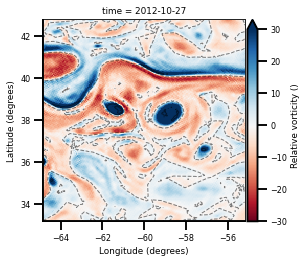} &
\includegraphics[trim={13mm 13mm 21.2mm 5mm},clip, width=3.2cm,height=3cm]{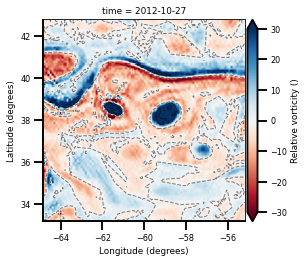} &
\includegraphics[trim={13mm 13mm 0 5mm},clip,width=4.0cm,height=3cm]{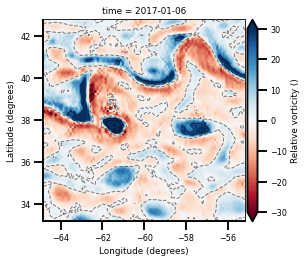} \\
\includegraphics[trim={0 0 19mm 5mm},clip, width=3.60cm,height=3.4cm]{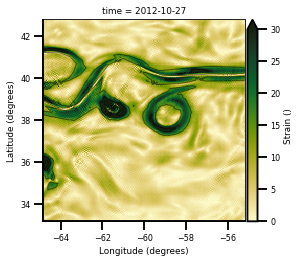} &
\includegraphics[trim={13mm 0 19mm 5mm},clip, width=3.2cm,height=3.4cm]{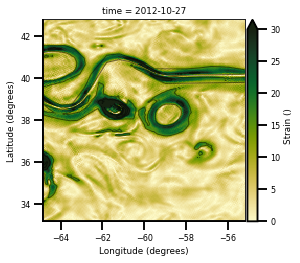} &
\includegraphics[trim={13mm 0 19mm 5mm},clip, width=3.2cm,height=3.4cm]{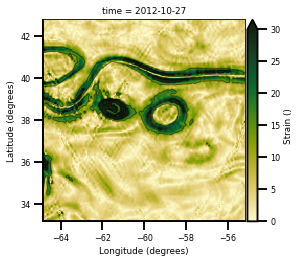} &
\includegraphics[trim={13mm 0 0 5mm},clip,width=4.0cm,height=3.4cm]{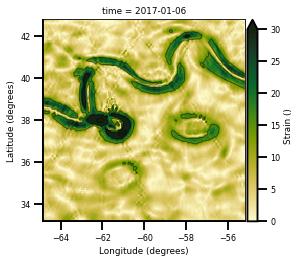} \\
(a) & (b) & (c) & (d)
\end{tabular}
\vspace{-3mm}
\caption{
Reconstructed quantities by the 4dVarNet method for each of the four tasks.
Each row showcases the following physical variables found in appendix~\ref{sec:physical_variables}: (a) Sea Surface Height, (b) Kinetic Energy, (c) Relative Vorticity, and (d) Strain. 
Each column showcase the reconstructed from the tasks (a) OSSE using only Nadir tracks: (b) OSSE using Nadir tracks and SWOT swath, (c) Multimodal using Nadir tracks and sea surface temperature, and (d) Reconstruction using real nadir altimetry tracks.}
\vspace{-5mm}
\label{fig:oceanbench_maps_4dvarnet}
\end{center}
\end{figure}

We use \texttt{OceanBench} to generate maps of relevant quantities from the 4DVarNet method~\cite{4DVARNETSWOT,4DVARNETSST}.
Figure~\ref{fig:oceanbench_maps_4dvarnet} showcases some demo maps for some key physical variables outlined in section~\ref{sec:physical_variables}.
We showcase the 4DVarNet method because it is the SOTA method that was applied to each of the data challenges.
We can see that the addition of more information, i.e. NADIR -> SWOT -> SST, results in maps look more similar to the NEMO simulation in the OSSE challenges.
It also produces sensible maps for the OSE challenge as well.

\begin{figure}[t!]
\small
\begin{center}
\setlength{\tabcolsep}{1pt}
\begin{tabular}{cccc}
\hspace{3mm} Task OSSE & 
 Task OSSE & 
\hspace{-10mm} Task OSSE & 
\hspace{-10mm}Task OSE \\
\hspace{3mm}  Nadir & 
 Nadir + SWOT & 
\hspace{-10mm} Nadir + SST & 
\hspace{-10mm}Nadir \\
\includegraphics[trim={0 0 0 0},clip, width=3.70cm,height=3.5cm]{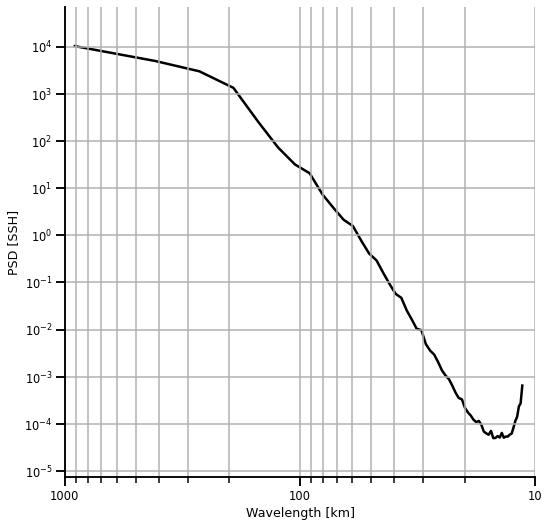} &
\includegraphics[trim={18mm 0 0 0},clip, width=3.3cm,height=3.5cm]{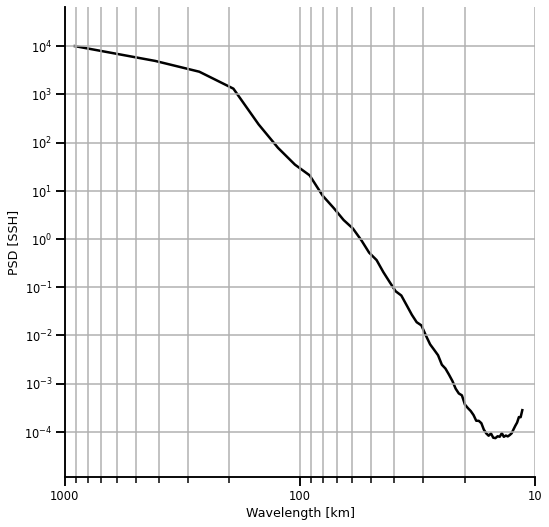} &
\hspace{-5mm}\includegraphics[trim={18mm 0 0 0},clip, width=3.3cm,height=3.5cm]{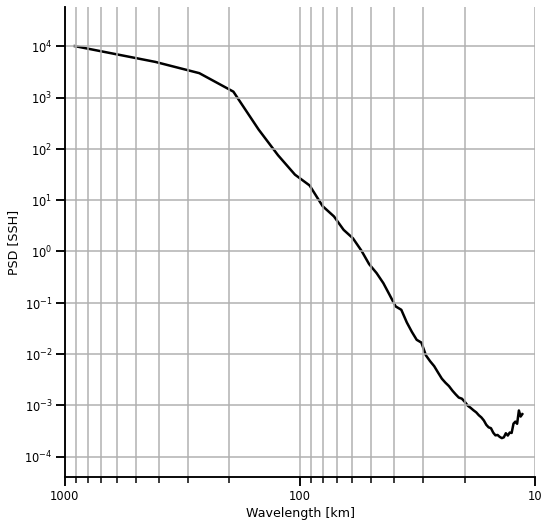} &
\hspace{-10mm}\includegraphics[trim={18mm 0 0 0},clip,width=3.3cm,height=3.5cm]{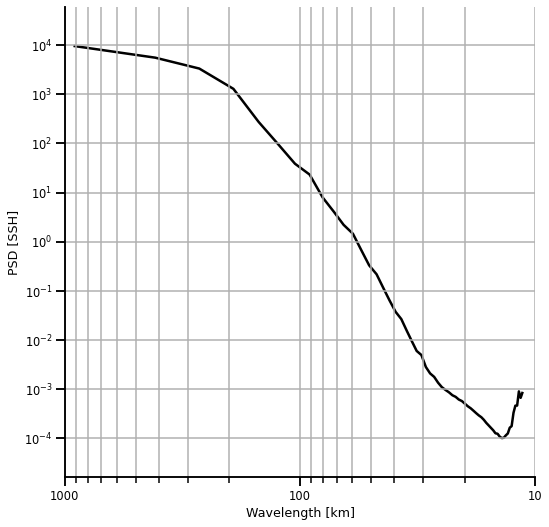} \\
\includegraphics[trim={0 0 0 0}, clip, width=3.70cm,height=3.5cm]{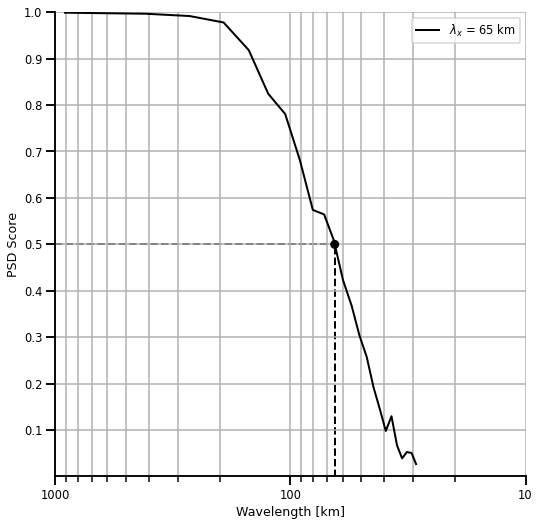} &
\hspace{1mm}\includegraphics[trim={18mm 0 0 0},clip, width=3.3cm,height=3.5cm]{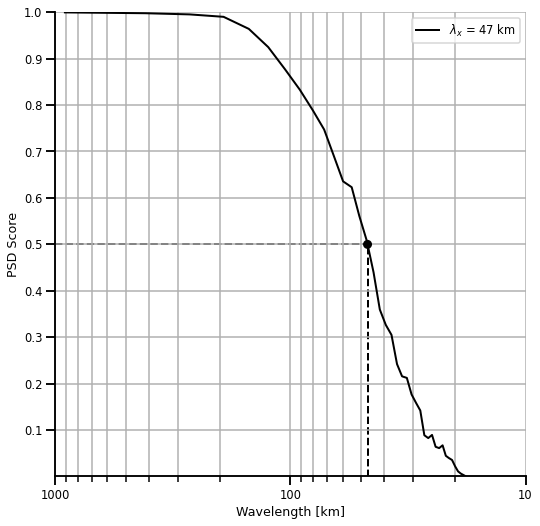} &
\hspace{-4mm}\includegraphics[trim={18mm 0 0 0},clip, width=3.3cm,height=3.5cm]{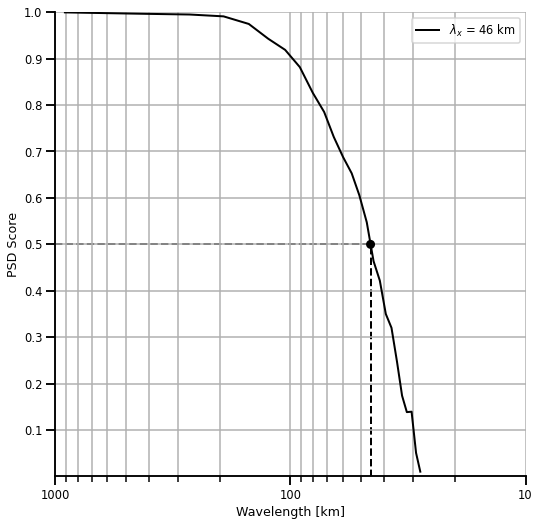} &
\hspace{-10mm}\includegraphics[trim={18mm 0 0 0},clip,width=3.3cm,height=3.5cm]{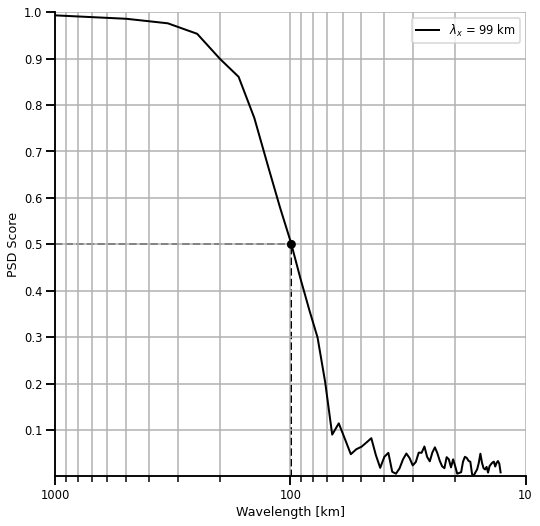} \\
\hspace{-4mm}\includegraphics[trim={0 0 23mm 0},clip, width=3.65cm,height=3.5cm]{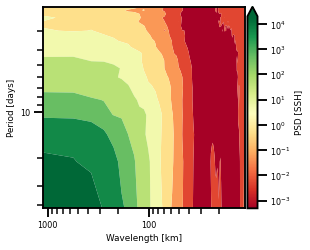} &
\includegraphics[trim={14mm 0 23mm 0},clip, width=3cm,height=3.5cm]{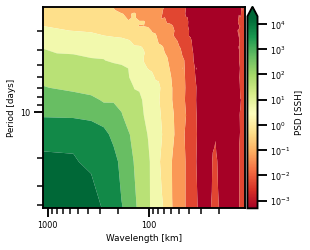} &
\hspace{-5mm}\includegraphics[trim={14mm 0 23mm 0},clip, width=3cm,height=3.5cm]{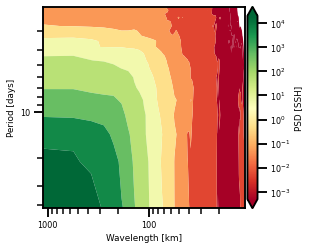} &
\hspace{-5mm}\includegraphics[trim={14mm 0 0 0},clip,width=3.8cm,height=3.5cm]{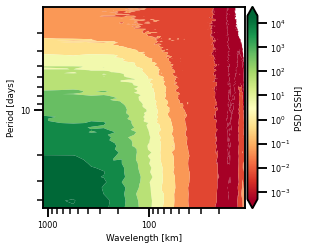} \\
\hspace{-4mm}\includegraphics[trim={0 0 23mm 0},clip, width=3.70cm,height=3.5cm]{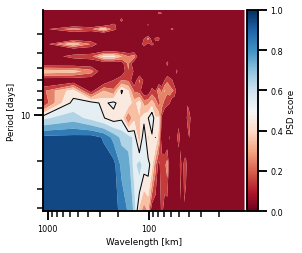} &
\hspace{-2mm}\includegraphics[trim={13mm 0 23mm 0},clip, width=3.1cm,height=3.5cm]{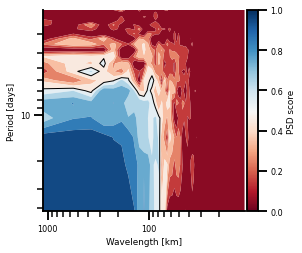} &
\hspace{1mm}\includegraphics[trim={13mm 0 0 0},clip, width=3.8cm,height=3.5cm]{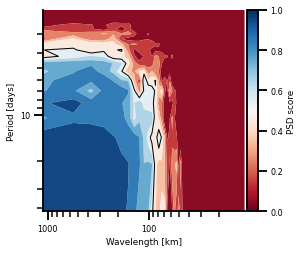} &
 \\
 \hspace{1mm} (a) & \hspace{-5mm} (b) & \hspace{-8mm}(c) & \hspace{-10mm}(d)
\end{tabular}
\vspace{-3mm}
\caption{
Power spectrum and associated scores of the 4dVarNet method for each of the four tasks.
The row display in order: (1) the isotropic PSD, (2) the spatial PSD score (using the isotropic PSD for the first three rows and along track PSD for the last row), (3) the space-time PSD, (4) The spacetime PSD score available only in OSSE task.  }

\vspace{-5mm}
\label{fig:oceanbench_psd_4dvarnet}
\end{center}
\end{figure}

\texttt{OceanBench} also generated figure~\ref{fig:oceanbench_psd_4dvarnet} which shows plots of the PSD and PSD scores of SSH for the different challenges.
Again, as we increase the efficacy of the observations via SWOT and allow for more external factors like the SST, we get an improvement in the isotropic and spacetime PSD scores.
In addition, we see that the PSD plots for the OSE task look very similar to the OSE challenges. 

Lastly, we used \texttt{OceanBench} to generate a leaderboard of metrics for a diverse set of algorithms where the maps were available online.
Table~\ref{tb:exp-results-mega} displays all of the key metrics outlined in section~\ref{sec:metrics} including the normalized RMSE and various spectral scores which are appropriate for the challenge.
We see that as the complexity of the method increases, the metrics improve. 
In addition, the methods that involve end-to-end learning perform the best overall, i.e. 4DVarNet.

\begin{table}[ht]
\caption{This table showcases all of the summary statistics for some methods for each of the data challenges listed in section~\ref{sec:data_challenges} and~\ref{sec:data_challenges_extended}. The summary statistics shown are the normalized RMSE and the effective resolution in the spectral domain. The spectral metrics for the effective resolution that were outlined in section~\ref{sec:metrics} are: i) $\lambda_a$ is the spatial score for the alongtrack PSD score, ii) $\lambda_r$ is the spatial score for the isotropic PSD, iii) $\lambda_x$ is the spatial score for space-time PSD score, and iv) $\lambda_t$ is the temporal score for the space-time PSD score.}

\label{tb:exp-results-mega}
\centering
\begin{tabular}{llcccccc}
 \toprule
\multirow{2}{*}{Experiment} &  \multirow{2}{*}{Algorithm} &   \multirow{2}{*}{nRMSE Score} &
\multicolumn{4}{c}{Effective Resolution} \\
& & & $\lambda_{a}$ [km] & $\lambda_{r}$ [km]   &  $\lambda_{\mathbf{x}}$ [km]  &   $\lambda_{t}$ [days]      \\ \midrule
OSSE NADIR     &  OI & 0.92 & - & 123 & 174 & 10.8 \\
OSSE NADIR     &  MIOST &  0.93 & - & 100 & 157 & 10.1 \\
OSSE NADIR     &  BFNQG & 0.93 & - & 88 & 139 & 10.4 \\
OSSE NADIR &  4DVarNet &  \textbf{0.94} & - & \textbf{65} & \textbf{117} & \textbf{7.7} \\
\midrule
OSSE SWOT     &  OI & 0.92 & - & 106 & 139 & 11.7 \\
OSSE SWOT     &  MIOST &  0.94 & - & 88 & 131 & 10.1 \\
OSSE SWOT     &  BFNQG & 0.94 & - & 64 & 118 & 36.5 \\
OSSE SWOT &  4DVarNet &  \textbf{0.96} & - & \textbf{47} & \textbf{77} & \textbf{5.6} \\
\midrule
OSSE SST     &  Musti & 0.95 & - & 46 & 138 & 4.1 \\
OSSE SST &  4DVarNet &  \textbf{0.96} & - & \textbf{46} & \textbf{87} & \textbf{3.7} \\
\midrule
OSE NADIR     &  OI & 0.88 & 151 & - &  - &  -\\
OSE NADIR     &  MIOST &  0.90 & 135 & - &  - &  -\\
OSE NADIR     &  BFNQG & 0.88 & 122 & - & - &  -\\
OSE NADIR &  ConvLSTM &  0.89 & 113 &- &  - &  -\\
OSE NADIR &  4DVarNet & \textbf{0.91} & \textbf{98} & - &  -  &  -\\
\bottomrule
\end{tabular}
\end{table}

\subsection{Datasets} \label{sec:datasets}

In Table~\ref{tb:datasets-mega}, we showcase all of the available datasets in our\footnote{Available at: \href{https://github.com/quentinf00/oceanbench-data-registry}{OceanBench Data Registry}} for the challenges listed in the above section. The license for the datasets listed in the registry are under the CCA 4.0 International License.

\begin{landscape}
\begin{table}[ht]
\caption{This table gives an extended overview of the datasets provided to complete the data challenges listed in~\ref{sec:data_challenges} and~\ref{sec:data_challenges_extended}. The OSSE SST and SSH are outputs from come from the free run NEMO model~\citep{NEMOAJAYI2020}. The OSSE NADIR and SWOT are pseudo-observations generated from the NEMO simulation. We provide the original simulated satellite tracks as well as a gridded version at the same resolution as the simulation. 
}
\label{tb:datasets-mega}
\centering
\begin{tabular}{lcclclcc}
 \toprule
     & OSSE SSH      & \multicolumn{2}{c}{OSSE SSH NADIR}                     & \multicolumn{2}{c}{OSSE SSH SWOT}                      & OSSE SST             & OSE SSH NADIR            \\ \midrule\midrule
Data Structure & Gridded              & AlongTrack           & \multicolumn{1}{c}{Gridded} & AlongTrack           & \multicolumn{1}{c}{Gridded} & Gridded              & AlongTrack           \\
     & \multicolumn{1}{l}{} & \multicolumn{1}{l}{} &                             & \multicolumn{1}{l}{} &  \\ \midrule
Source     & 
NEMO~\citep{NEMOAJAYI2020} &
NEMO~\citep{NEMOAJAYI2020} &
NEMO~\citep{NEMOAJAYI2020} &
NEMO~\citep{NEMOAJAYI2020} & 
NEMO~\citep{NEMOAJAYI2020} &
NEMO~\citep{NEMOAJAYI2020}
& Altimetry~\citep{MDSALONGTRACK} \\
Region & 
GulfStream & GulfStream & GulfStream & GulfStream &
GulfStream & GulfStream & GulfStream
\\
Domain Size [degrees] &
$10\times 10^\circ$ &
$10\times 10^\circ$ &
$10\times 10^\circ$ &
$10\times 10^\circ$ &
$10\times 10^\circ$ &
$10\times 10^\circ$ &
$10\times 10^\circ$ \\
Domain Size [km] &
$1,100\times 1,100$ &
$1,100\times 1,100$ &
$1,100\times 1,100$ &
$1,100\times 1,100$ &
$1,100\times 1,100$ &
$1,100\times 1,100$ &
$1,100\times 1,100$ \\
Longitude Extent &
$[-65^\circ, -55^\circ]$ & 
$[-65^\circ, -55^\circ]$ & 
$[-65^\circ, -55^\circ]$ & 
$[-65^\circ, -55^\circ]$ & 
$[-65^\circ, -55^\circ]$ &
$[-65^\circ, -55^\circ]$ &
$[-65^\circ, -55^\circ]$ \\
Latitude Extent &
$[33^\circ, 43^\circ]$ &
$[33^\circ, 43^\circ]$ &
$[33^\circ, 43^\circ]$ &
$[33^\circ, 43^\circ]$ &
$[33^\circ, 43^\circ]$ &
$[33^\circ, 43^\circ]$ &
$[33^\circ, 43^\circ]$ \\
Resolution [degrees] &
$0.05^\circ\times 0.05^\circ$ &
N/A &
$0.05^\circ\times 0.05^\circ$ &
N/A &
$0.05^\circ\times 0.05^\circ$ &
$0.05^\circ\times 0.05^\circ$ &
N/A \\
Resolution [km] &
$5.5\times 5.5$ &
$6$ &
$5.5\times 5.5$ &
$6$ &
$5.5\times 5.5$ &
$5.5\times 5.5$ &
$7$ \\
Grid Size &
$200\times 200$ & 
N/A &
$200\times 200$ & 
N/A &
$200\times 200$ & 
$200\times 200$ & 
N/A \\
Num. Datapoints &
$\sim$14.6M & 
$\sim$205K & 
$\sim$14.6M & 
$\sim$955K & 
$\sim$14.6M & 
$\sim$14.6M & 
$\sim$1.79M \\ \midrule
Period Start & 
2012-10-01 & 2012-10-01 & 2012-10-01 & 2012-10-01 & 
2012-10-01 & 2012-10-01 & 2016-12-01 \\
Period End & 
2013-09-30 & 2013-09-30 & 2013-09-30 & 2013-09-30 & 
2013-09-30 & 2013-09-30 & 2018-01-31 \\
Frequency  & 
Daily & 1 Hz  & Daily & 1 Hz  & Daily & Daily & 1 Hz \\ 
Period Length & 365 Days & 365 Days & 365 Days &
365 Days & 365 Days & 365 Days & 427 Days \\
\midrule
Evaluation Start & 
2012-10-22 & 2012-10-22 & 2012-10-22 & 2012-10-22 & 
2012-10-22 & 2012-10-22 & 2017-01-01 \\
Evaluation End & 
2012-12-02 & 2012-12-02 & 2012-12-02 & 2012-12-02 & 
2012-12-02 & 2012-12-02 & 2017-12-31 \\ 
Evaluation Length & 45 Days & 45 Days & 45 Days &
45 Days & 45 Days & 45 Days & 365 Days \\
\bottomrule
\end{tabular}
\end{table}
\end{landscape}
\newpage
\section{Physical Variables} \label{sec:physical_variables}

As alluded to in the main body of the paper, we have access to many physical quantities which can be derived from  sea surface height. 
This gives us a way to analyze how effective and trustworthy are our reconstructions. 
Many machine learning methods are unconstrained so they may provide solutions that are physically inconsistent and visualizing the field is a very easy eye test to assess the validity. 
In addition to post analysis, one could include some of these derived quantities maybe useful as additional inputs to the system and/or constraints to the loss function. 
Recall the spatiotemporal coordinates from equation~\ref{eq:spatiotemporal_coords}, 
we use the same coordinates for the subsequent physical quantities. \textbf{Sea Surface Height} is the deviation of the height of the ocean surface from the geoid of the Earth. We can define it as:
\begin{align}
	\text{Sea Surface Height }[m]:&& \quad
 \eta &= \boldsymbol{\eta}(\mathbf{x},t)&& \quad \Omega\times \mathcal{T}\rightarrow\mathbb{R} \label{eq:ssh}
\end{align}
This quantity is the actual value that is given from the satellite altimeters and is presented in the products for SSH maps~\cite{DUACS}. An example can be seen in the first row of figure~\ref{fig:oceanbench_maps_4dvarnet}.

\textbf{Sea Surface Anomaly} is the anomaly wrt to the spatial mean which is defined by
\begin{align}
	\text{Sea Level Anomaly }[m]:&& \quad
 \bar{\eta} &= \boldsymbol{\eta}(\mathbf{x},t) - \bar{\eta}(t) &&
 \quad \Omega\times \mathcal{T}\rightarrow\mathbb{R} \label{eq:sla}
\end{align}
where $\bar{\eta}(t)$ is the spatial average of the field at each time step.  
An example can be seen in the first row of figure~\ref{fig:oceanbench_maps}.

Another important quantity is the \textbf{geostrophic velocities} in the zonal and meridional directions. This is given by
\begin{align}
	\text{Zonal Velocity}[ms^{-2}]:&& \quad
 u &= -\frac{g}{f_0}\frac{\partial \eta}{\partial y} &&
 \quad \Omega\times \mathcal{T}\rightarrow\mathbb{R} \label{eq:u_vel} \\
	\text{Meridional Velocity}[ms^{-2}]:&& \quad
 v &= \frac{g}{f_0}\frac{\partial \eta}{\partial x} &&
 \quad \Omega\times \mathcal{T}\rightarrow\mathbb{R} \label{eq:v_vel}
\end{align}
where $g$ is the gravitational constant and $f_0$ is the mean Coriolis parameter. These quantities are important as they can be an related to the sea surface current. The geostrophic assumption is a very strong assumption however it can still be an important indicator variable. The \textbf{kinetic energy} is a way to summarize the (geostrophic) velocities as the total energy of the system. This is given by
\begin{equation} \label{eq:kineticenergy}
    KE = \frac{1}{2}\left(u^2 + v^2\right)
\end{equation}
An example can be seen in the second row of figure~\ref{fig:oceanbench_maps_4dvarnet}.

Another very important quantity is the \textit{vorticity} which measures the spin and rotation of a fluid. In geophysical fluid dynamics, we use the \textbf{relative vorticity} which is the vorticity observed within at rotating frame.
This is given by
\begin{equation} \label{eq:relvorticity}
    \zeta = \frac{\partial v}{\partial x} - \frac{\partial u}{\partial y}
\end{equation}
An example can be seen in the third row of figure~\ref{fig:oceanbench_maps_4dvarnet}.



We can also use the \textbf{Enstrophy} to summarize the relative voriticty to measure the total contribution which is given by
\begin{equation} \label{eq:enstrophy}
    E = \frac{1}{2}\zeta^2
\end{equation}

The \textbf{Strain} is a measure of deformation of a fluid flow.

\begin{equation} \label{eq:strain}
    \sigma = \sqrt{\sigma_n^2 + \sigma_s^2}
\end{equation}

where $\sigma_n$ is the shear strain (aka the shearing deformation) and $\sigma_s$ is the normal strain (aka stretching deformation). An example can be seen in the fourth row of figure~\ref{fig:oceanbench_maps_4dvarnet}.

The \textbf{Okubo-Weiss Parameter} is high-order quantity which is a linear combination of the strain and the relative vorticity.

\begin{equation} \label{eq:okuboweiss}
    \sigma_{ow} = \sigma_n^2 + \sigma_s^2 - \zeta^2
\end{equation}

This quantity is often used as a threshold for determining the location of Eddies in sea surface height and sea surface current fields~\cite{OKUBO, WEISS, OKUBOWEISS}.

\newpage
\section{Metrics} \label{sec:metrics}

There are many metrics that are standard within the ML community but unconvincing for many parts the geoscience community. 
Specifically, many of these standard scores do not capture the important optimization criteria in the scientific machine learning tasks.
However, there is not consensus within domain-specific communities about the perfect metric which captures every aspect we are interested.
Therefore, we should have a variety of scores from different perspectives to really assess the pros and cons of each method we wish to evaluate thoroughly. 
Below, we outline two sets of scores we use within this framework: skill scores and spectral scores.

\subsection{Skill Scores}

We classify one set of metrics as \textit{skill scores}. 
These are globally averaged metrics which tend to operate within the real space.
Some examples include the root mean squared error (RMSE), the normalized root mean squared (nRMSE) error, and the nRMSE score.
The RMSE metric can also be calculated w.r.t. the spatial domain, temporal domain or both. 
For example, figure~\ref{fig:oceanbench_psd} showcases the nRMSE score calculated only on the spatial domain and visualized for each time step.
\begin{align}
    \text{RMSE}: &&\text{RMSE}(\eta,\hat{\eta}) &= ||\eta - \hat{\eta}||_2 \label{eq:RMSE}\\
    \text{nRMSE}: &&\text{nRMSE}(\eta,\hat{\eta}) &= \frac{\text{RMSE}(\eta,\hat{\eta})}{||\eta||_2} \label{eq:nRMSE} \\
    \text{nRMSE}_{\text{score}}: &&\text{nRMSE}_{\text{score}}(\eta,\hat{\eta}) &= 1 - \text{nRMSE}(\eta,\hat{\eta})
    \label{eq:nRMSE_score}
\end{align}
However, we are not limited to just the standard MSE metrics.
We can easily incorporate more higher-order statistics like the Centered Kernel Alignment (CKA)~\cite{METRICSCKA} or information theory metrics like mutual information (MI)~\cite{METRICSITRBIG,METRICSITRBIG2}.
In addition, we could also utilize the same metrics in the frequency domain as is done in~\citep{PDEBench}.

\subsection{Spectral Scores}

Another class of scores that we use in \texttt{OceanBench} are the \textit{spectral scores}. These scores are calculated within the spectral space via the wavenumber power spectral density (PSD). 
This provides a spatial-scale-dependent metric which is useful for identifying the largest and smallest scales that were resolved by the reconstruction map. 
In general, we use these to measure the expected energy at different spatiotemporal scales and we can also construct custom score functions which gives us a summary statistic for how well we reconstructed certain scales.
\begin{align}
    \text{PSD}: &&\text{PSD}(\eta) &= \sum_{k_{min}}^{k_{max}}\|\mathcal{\mathcal{F}(\eta)}\|^2\label{psd}\\
    \text{PSD}_{score}: &&\text{PSD}_{score}(\eta,\hat{\eta}) &= 1 - \frac{\text{PSD}(\eta - \hat{\eta})}{\text{PSD}(\eta)} \label{eq:psd_score}
\end{align}
where $\mathcal{F}$ is the Fast Fourier Transformation (FFT). 
In our application, there are various ways to construct the PSD which depend on the FFT transformation.
We denote the \textit{space-time PSD} as $\lambda_\mathbf{x}$ which does the 2D FFT in the longitude and time direction, then takes the average over the latitude.
We denote the \textit{space-time PSD} as $\lambda_\mathbf{t}$ which does the 2D FFT in the longitude and latitude direction, then takes the average over the time.
We denote the \textit{isotropic PSD} as $\lambda_r$ which assumes a radial relationship in the spatial domain and then averages over the temporal domain.
Lastly, we denote the standard PSD score as $\lambda_a$ which is the 1D FFT over a prescribed distance along the satellite track; this is what is done for the OSE NADIR experiment.
We recognize that the FFT configurations are limited due to their global treatment of the spectral domain and we need more specialized metrics to handle the local scales.
This opens the door to new metrics that handle such cases such as the Wavelet transformation~\cite{METRICSWAVELET}.

\newpage
\section{Use Case II: Hydra Recipes} \label{sec:hydra_recipes}

This framework has drastically reduced the overhead for the ML researcher while also enhancing the reprducibility and replicability of the preprocessing steps. In this section we showcase a few examples for how one can use oceanbench in conjunction with hydra to provide recipes for some standard processes.

\subsection{Task Recipe} \label{sec:hydra_recipe_task}

In this example, we showcase how we define an interpolation task for the OSE NADIR data challenge. 
We need to state the list of datasets available and specify which datasets are to be using for training and testings.
We also specify the spatial region we would like to train on and the train-test period.
There are a few simple changes one could do here to extend this task provided that one has uploaded standardized data that follows our set conventions.
For example, for this interpolation task, the test period is a complete subset of the train period but one could imagine a forecasting task whereby the test period is at a completely different time period.
Similarly, for this task, the train-test domain is the same but we could easily change the region of interest to see how the models perform in a completely different domain.

\begin{listing}[ht!]
\begin{minted}[frame=lines]{yaml}
#@package _global_.task
outputs:
    # name of data challenge
    name: DC2021 OSE Gulfstream
    # list of datasets and locations
    data:
      train: # train data list
        alg: ${....data.outputs.alg}
        h2g: ${....data.outputs.h2g}
        j2g: ${....data.outputs.j2g}
        j2n: ${....data.outputs.j2n}
        j3: ${....data.outputs.j3}
        s3a: ${....data.outputs.s3a}
      test: # test data list
        c2: ${....data.outputs.c2}
    # spatial region specification
    domain: {lat: [33, 43], lon: [-65, -55]}
    # temporal period specification
    splits: {
        test: ['2017-01-01', '2017-12-31'], 
        train: ['2016-12-01', '2018-01-31']
    }
\end{minted}
\label{hydraconfig:task}
\caption{This is a \texttt{.yaml} which showcases how we can communicate with \texttt{Hydra} framework to list a predefined set of specifications for the spatial region and the temporal period. This is an interpolation task for the OSE NADIR data challenge listed in section~\ref{sec:ose_nadir}.}
\end{listing}

\newpage
\subsection{GeoProcessing Recipe} \label{sec:hydra_geoprocess_task}

In this example, we showcase how one can pipe a sequential transformation through the hydra framework. In this example, we open the dataset, validate the coordinates to comply to our standards, select the region of interest, subset the data, regrid the alongtrack data to a uniform grid, and save the data to a netcdf file. See the listing~\ref{hydraconfig:geoprocess} for more information.

\begin{listing}[ht!]
\begin{minted}[frame=lines]{yaml}
# Target Function to initialize
_target_: "oceanbench._src.dataset.pipe"
# netcdf file to be loaded
inp: "${data_directory}/nadir_tracks.nc"
# sequential transformations to be applied
fns:
    # Load Dataset
    - {_target_: "xarray.open_dataset", _partial_: True}
    # Validate LatLonTime Coordinates
    - {_target_: "oceanbench.validate_latlon", _partial_: True}
    - {_target_: "oceanbench.validate_time", _partial_: True}
    # Select Specific Region (Spatial | Temporal)
    - {_target_: "xarray.Dataset.sel", args: ${domain}, _partial_: True}
    # Take Subset of Data
    - {_target_: "oceanbench.subset", num_samples: 1500, _partial_: True}
    # Regridding (AlongTrack -> Uniform Grid)
    - {
        _target_: "oceanbench.regrid", 
        target_grid: ${grid.high_res}, 
        _partial_: True
      }
    # Save Dataset
    - {
        _target_: "xarray.Dataset.to_netcdf", 
        save_name: "demo.nc", 
        _partial_: True
      }
\end{minted}
\label{hydraconfig:geoprocess}
\caption{This is a \texttt{.yaml} which showcases how we can communicate with \texttt{Hydra} framework to list a predefined set of transformations to be \textit{piped} through sequentiall. In this example, we showcase some standard pre-processing strategies to be saved to another netcdf file.}
\end{listing}

\newpage
\subsection{Evaluation Recipe - OSSE} \label{sec:hydra_evaluation_task}

In this example, we showcase how one can use hydra to do the evaluation procedure. This is the same evaluation procedure that is used to evaluate the effectiveness of the OSSE NADIR experiment. From code snippet~\ref{hydraconfig:geoprocess}, we see that we choose which target function to initialize and we choose the data directory where the \texttt{.netcdf} file is located. Then, we pipe some transformations for the \texttt{.netcdf} file: 1) validate the spatiotemporal coordinates, 2) we select the evaluation region, 3) we regrid it to the target get, 4) we fill in the nans with a Gauss-Seidel procedure, 5) we rescale the coordinates to be in meters and days, and 6) we perform the isotropic power spectrum transformation to get the effective resolution outlined in section~\ref{sec:metrics}.

\begin{listing}[ht!]
\begin{minted}[frame=lines]{yaml}
# Target Function to initialize
_target_: "oceanbench._src.dataset.pipe"
# netcdf file to be loaded
inp: "${data_directory}/ml_result.nc"
# sequential transformations to be applied
fns:
    # Load Dataset
    - {_target_: "xarray.open_dataset", _partial_: True}
    # Validate LatLonTime Coordinates
    - {_target_: "oceanbench.validate_latlon", _partial_: True}
    - {_target_: "oceanbench.validate_time", _partial_: True}
    # Select Specific Region (Spatial | Temporal)
    - {_target_: "xarray.Dataset.sel", args: ${domain}, _partial_: True}
    # Regridding (Uniform Grid -> Uniform Grid)
    - {_target_: "oceanbench.regrid", 
       target_grid: ${grid.reference}, _partial_: True}
    # Fill NANS (around the corners)
    - {_target_: "oceanbench.fill_nans", 
       method: "gauss_seidel", _partial_: True}
    # Coordinate Change (degree -> meters, ns -> days)
    - {_target_: "oceanbench.latlon_deg2m", _partial_: True}
    - {_target_: "oceanbench.time_rescale", 
       freq: 1, unit: "days", _partial_: True}
    # Calculate Isotropic Power Spectrum
    - {_target_: "oceanbench.power_spectrum_isotropic", 
       reference: ${grid.reference}, _partial_: True}
    # Calculate Resolved Spatial Scale
    - {_target_: "oceanbench.resolved_scale", _partial_: True}
    # Save Dataset
    - {_target_: "xarray.Dataset.to_netcdf", 
       save_name: "ml_result_psd.nc", _partial_: True}
\end{minted}

\label{hydraconfig:evaluation}
\caption{This is a \texttt{.yaml} which showcases how we can communicate with \texttt{Hydra} framework to list a predefined set of transformations to be \textit{piped} through sequential. In this example, we showcase some standard pre-processing strategies to be saved to another netcdf file.}
\end{listing}

\newpage
\section{Use Case III: XRPatcher} \label{sec:xrpatcher}

There are many usecases for the \texttt{XRPatcher}. For example, we can do 1D Time chunking, 2D Spatial-Temporal Patches, or 3D Spatial-Temporal Cubes.

\begin{listing}[h!]
\begin{minted}[frame=lines]{python}
import xarray as xr
import torch
import itertools
from oceanbench import XRPatcher
# Easy Integration with PyTorch Datasets (and DataLoaders)
class XRTorchDataset(torch.utils.data.Dataset):
    def __init__(self, batcher: XRPatcher, item_postpro=None):
        self.batcher = batcher
        self.postpro = item_postpro
    def __getitem__(self, idx: int) -> torch.Tensor:
        item = self.batcher[idx].load().values
        if self.postpro:
            item = self.postpro(item)
        return item
    def reconstruct_from_batches(
            self, batches: list(torch.Tensor), **rec_kws
        ) -> xr.Dataset:
        return self.batcher.reconstruct(
            [*itertools.chain(*batches)], **rec_kws
        )
    def __len__(self) -> int:
        return len(self.batcher)
# load demo dataset
data = xr.tutorial.load_dataset("eraint_uvz")
# Instantiate the patching logic for training
patches = dict(longitude=30, latitude=30)
train_patcher = XRPatcher(
    da=data,
    patches=patches,
    strides=patches,        # No Overlap
    check_full_scan=True    # check no extra dimensions
)
# Instantiate the patching logic for testing
patches = dict(longitude=30, latitude=30)
strides = dict(longitude=5, latitude=5)
test_patcher = XRPatcher(
    da=data,
    patches=patches,
    strides=strides,        # Overlap
    check_full_scan=True    # check no extra dimensions
)
# instantiate PyTorch DataSet
train_ds = XRTorchDataset(train_patcher, item_postpro=TrainingItem._make)
test_ds = XRTorchDataset(test_patcher, item_postpro=TrainingItem._make)
# instantiate PyTorch DataLoader
train_dl = torch.utils.data.DataLoader(train_ds, batch_size=4, shuffle=False)
test_dl = torch.utils.data.DataLoader(test_ds, batch_size=4, shuffle=False)
\end{minted}
\label{listing:xrpatcher}
\caption{This is a snippet showcasing how we can easily integrate PyTorch Datasets within the \texttt{XRPatcher} framework without much overhead. Here we define a custom PyTorch Dataset to handle the \texttt{XRPatcher}. We load an arbitrary dataset with \texttt{xarray}, then we instantiate the \texttt{XRPatcher} with some patching logic, then we instantiate the PyTorch dataset and dataloader as per usual.}
\end{listing}

\newpage
\section{Additional Tasks}\label{sec:other_tasks}

In the main paper, we thoroughly outlined the interpolation task to showcase how \texttt{OceanBench} can be used to create automated pipelines for processing and evaluation procedures.
However, there are many other additional tasks that can make use of the \texttt{OceanBench} features. 

\textbf{Denoising}. A simpler problem for interpolation tasks is the denoising problem~\cite{DENOISESURVEY,DENOISESURVEY2}.
The SSH and SST measurements we obtain have inherent noise from the sensors.
A key problem is to calibrate the observations by separating the known noise patterns and the true signal.
There has already been a lot of work from the ML side ranging from amortized predictions~\cite{DENOISESWOT} to end-to-end learning schemes~\cite{DENOISESWOT2}.
Much of this work has been facilitated by the \textit{Ocean-Data-Challenge} group which have a few data challenges related to the denoising problem.
Just like \texttt{OceanBench} was able to create reproducible pipelines from the SSH interpolation challenge listed in section~\ref{sec:data_challenges}, we also believe that one could extend the denoising challenge in the same manner.

\textbf{Forecasting}. This is a special form of extrapolation whereby the temporal domain of the state variable is sufficiently outside of the domain of the observation domain. 
Many previous benchmarking suites already look at forecasting for weather~\cite{weatherbench} and climate~\cite{ClimateBench}.
However, in oceanography, it is also advantageous to do forecasting for problems involving currents~\cite{MLSSC,4DVarNetSSC} and eddies~\cite{OCEANEddyTracking,OKUBO,OKUBOWEISS}.
The \texttt{xrpatcher} will work out of the box for forecasting problems and contributions can be made to \texttt{OceanBench} to include some specific metrics for forecasting as were outlined in~\cite{weatherbench,ClimateBench,ENS10Bench}.

\textbf{Proxy Variables}. There are many other control variables that one could use to improve the interpolation or extrapolation task.
We mentioned SST in section~\ref{sec:data_challenges} because it is the most abundant observations available.
However, there are other important observed variables which could be useful, e.g. Ocean colour, Bigeochemical parameters, and atmospheric variables.
In many other downstream applications, the oceanography community often uses SSH and SST as proxy variables to predict important quantities related to the carbon uptake, e.g. SOCAT~\cite{SOCAT}.
It would be straightforward to include a specific variable (and the associated preprocessing operations) into \texttt{OceanBench}.

\textbf{Dimension Reduction}. We often have very resolution spatiotemporal fields.
which poses a very big challenge for learning due to the high correlations exhibited by spatiotemporal data and high dimensionality.
A workaround for this is to learn a latent representation which retains as much relevant information as possible for the given task.
In the ocean sciences, this is known as \textit{Reduced Order Modeling} (ROM) or more generally dimensionality reduction which has been frequently used for adaptive meshes for physical models~\cite{NEMOEOF}.
This could be used for pretraining fields to latent embeddings which could be useful for downstream tasks like anomaly detection~\cite{SSTFLOWANOMALY}.

\textbf{Surrogate Modeling}. 
Physical model simulations are very expensive and ML has played a role in learning surrogate models to descrease the computational intensity~\cite{ML4OCN,MLCLOSURE}.
We have a decently long spatiotemporal field over a region of interest which could be used to learn a surrogate model to mimic the dynamics of that region.
This is also very useful for hybrid schemes whereby we have parameterizations to account for processes that are missing from low resolution simulations.~\cite{MLOCNPARAMETERIZATION,MLOCNPARAMETERIZATION2, MLOCNPARAMETERIZATION3, MLOCNPARAMETERIZATION4}.

\newpage
\section{Machine Learning Method Ontology} \label{sec:ml_ontology}

Although this paper does not focus on the explicit methods used for SSH interpolation, we would like to give a readers a brief overview of some of the most popular methods in the literature.

\subsection{Coordinate-Based methods}

These methods learn a direct mapping between the coordinate vectors to the scalar or vector values. 
\begin{align}
    \boldsymbol{y}_{obs} &= \boldsymbol{f}(\mathbf{x},t;\boldsymbol{\theta})+\boldsymbol{\epsilon}(\mathbf{x},t)
\end{align}
This is better known as \textit{functa}~\cite{FUNCTA} which parameterizes the field directly as a model.

\textbf{Functional}. Optimal Interpolation (OI) is the most common method used for many of the operational methods~\cite{DUACS}. It is a non-parametric, functional method which is built upon covariance and precision matrices. In the machine learning community, these methods are known as Gaussian Process~\cite{GPsBIGDATA} and in the geostatistics community, this is known as Kriging~\cite{KRIGINGREVIEW}.

\textbf{Basis Function}. This is an easy simplification to the functional by introducing parametric basis functions. In particular, the MIOST~\cite{MIOST} algorithm will be adopted in the new operational products for SSH interpolation. It is a custom basis function based on Wavelet analysis which is scale-aware and scalable.

\textbf{Neural Fields}. Neural fields (NerFs) are a very popular set of methods that use neural networks to effectively learn the basis function through a composition of weights, biases and activations~\cite{NERFSSSH}.
Furthermore, one can add physics-informed constraints to the loss function which mirror that of a PDE~\cite{PINNS}.
In many cases, especially with many auxillary inputs, we don't have access the PDE so one fit a NN directly to the observations with a fully connected neural network~\cite{SOCAT}.

\subsection{Grid-Based Methods}

In practice, we often consider the field at a specific discretized setting like a uniform grid or mesh. 
This is because we typically operate on and store these fields as multi-dimensional arrays which are only defined on a subspace of the entire continuous domain. 
We denote a discretized spatial representation as $\boldsymbol{\Omega}_g\subset\mathbb{R}^{N_s}$. 
We can simplify this notation by including the domain within the operator. So equation~\ref{eq:interp_problem} like so:
\begin{equation}\label{eq:interp_problem_discretized}
    \boldsymbol{\eta}(\boldsymbol{\Omega}_{obs},t ) = \mathcal{H}\left(\boldsymbol{\eta}(\boldsymbol{\Omega}_g,t), t, \boldsymbol{\mu},  \boldsymbol{\varepsilon} \right) 
\end{equation} 
In this equation, $\mathcal{H}$ is the observation operator that transforms the field from the full discretized domain, $\boldsymbol{\Omega}_g$, to the observation domain, $\boldsymbol{\Omega}_{obs}\subset\mathbb{R}^{N_{obs}}$.

\textbf{Direct Methods}. 
These methods take the noisy, incomplete observations and directly feed it to a model that returns the full reconstructed field.
They typically involve training a convolutional neural network or recurrent neural network on pairs of corrupted observations to learn the reconstruction~\cite{SuperResSurvey,IMAGE2IMAGETRANSLATION, IMAGE2IMAGETRANSLATION2}.
This has seem some sucess in applications related to SSH interpolation~\cite{SSHInterpUNet,SSHInterpConvLSTM, SSHInterpAttention}.

\textbf{Traditional Data Assimilation.}
There are many traditional methods that are rooted in data assimilation~\cite{DAGEOSCIENCE}.
For example, the GLORYS~\cite{GLORYS12} method propagates the physical model forwards in time and then \textit{updates} the state based on observations periodically.
A simpler approach is to use a nudging scheme coupled with a simpler physical model~\cite{BFNQG}.

\textbf{End-to-End Learning}. These methods try to solve the problem by learning and end-to-end scheme to solve the model inversion problem.
This is very similar to implicit methods that define a cost function to minimize instead of a minimizing the parameters of a prior model.
Plug-in-Play priors are a popular class of methods that pre-train priors on auxillary observations and then use the prior in the inversion scheme~\cite{DEEPUNFOLDING}.
This has seen a lot of success in SSH interpolation~\cite{4DVARNETSWOT,4DVARNETSST,4DVarNetSSC}.

\newpage
\section{Limitations} \label{sec:appendix_limitations}

\subsection{Framework Limitations}

While we have advertised \texttt{OceanBench} as a unifying framework that provides standardized processing steps that comply with domain-expert standards, we also highlight some potential limitations that could hinder its adoption for the wider community.

\textbf{Data Serving}. We provide a few datasets but we omit some of the original simulations. We found that the original simulations are terabytes/petabytes of data which becomes infeasible for most modest users (even with adequate CPU resources).  
This is very big problem and if we want to have a bigger impact, we may need to do more close collaborations with specified platforms like the Marine Data Store~\citep{MDSOCEANPHYSICS,MDSBIOGEOCHEMICAL,MDSOCEANPHYSICSENS,MDSINSITU,MDSWAVES,MDSALONGTRACK,MDSSSH} or the Climate Data Store~\citep{CDSREANALYSISSST,CDSOBSSST,CDSOBSOC,CDSOBSSSTENS}. Furthermore, there are many people that will not be able to do a lot of heavy duty research which indirectly favours institutions with adequate resources and marginalizing others. 
This is also problematic as those communities tend to be the ones who need the most support from the products of such frameworks.
We hope that leaving this open-source at least ensure that the knowledge is public.

\textbf{Framework Dependence}. The user has to "buy-into" the \texttt{hydra} framework to really take advantage of \texttt{OceanBench}. This adds a layer of abstraction and a new tool to learn. 
However, we designed the project so that high level usage does not require in-depth knowledge of the framework. 
In addition, we hope that, despite the complexity of project, users will appreciate the flexibility and extensibility of this framework.

\textbf{Lack of Metrics}. We do not provide the most exhaustive list of metrics available with the ocean community. In fact, we also believe that many of these metrics are often poor and do not effectively assess the goodness of our reconstructions. 
However, we do provide a platform that will hopefully be useful and easy to implement new and improved metrics.
Furthermore, having a wide range of metrics that are trusted across communities may help to improve the overall assessment of the different model performances~\cite{METRICSAVERAGE}.

\textbf{Limited ML Scope}. 
The framework does not support nor promote any machine learning methods and we lack any indication of comparing ML training and inference performance. 
However, we argue that a benchmark framework will allow us to effectively compare whichever ML methods are demonstratively the best which is a necessary preliminary step which offers users more flexibility in the long-run.

\textbf{Broad Oceans Application Scope}. 
We have targeted a broad ocean-application scope of state estimation.
However, there may be more urgent applications such as maritime monitoring, object tracking, and general ocean health.
However, we feel that many downstream applications require high-quality maps.
In addition, those downstream applications tend to be very complicated and are not always straightforward to apply ML under those instances.

\textbf{Full Pipeline Transparency}. We use a lot of different \texttt{xarray}-specific packages which have different design principles, assumptions and implementations. This may give the users an illusion of simplicity and transparency to real-world use. However, there are many underlying assumptions within each of the packages that may occlude a lot of design decisions.
Despite this limitation, we believe that being transparent about the processing steps and being consistent with the evaluation procedure will be beneficial for the ML research community.

\textbf{Scalability}. Scaling this to many terabytes or petabytes of data is easily the biggest limitation of the framework. In addition, we have only showcased demonstrations for 2D+T fields which are much less expensive than 3D+T fields.

\textbf{Deployability}. MLOPs has many wheels and it is not easy to integrate into existing systems. We offer no solutions to this. 
However, we believe that our framework is fully transparent in the assumptions and use cases which will facilitate some adoption into operational systems where they can further modify it for their use cases (see the evolution of \texttt{WeatherBench} and \texttt{ClimateBench}).

\textbf{Visualization Tools}.
We do not incorporate a high quality visualization tool that allows users to do pre- and post-analysis at a large scale. 
We do provide some simple visualization steps that are ML-relevant (see the GitHub repo) but it is very limited to ML standards.
One solution is to interface our pipeline with the source of many ocean datasets, e.g. Climate Data Store~\citep{CDSREANALYSISSST} or Marine Data Store~\citep{MDSOCEANPHYSICS}, then we can offset this task to them where they can offer better quality visualization tools.

\newpage
\subsection{Data Challenge Limitations}

We have showcased the SSH interpolation edition as a data challenge which could be helpful for real applications. 
However, in section~\ref{sec:problem_scope} we alluded to the greater task of general ocean state estimation which is more pertinent to the ocean sciences yet we don't address this directly with our data challenges.
Furthermore, we claim that the data challenges presented will help the ocean community with using ML for SSH interpolation.
Below, we outline some limitations which address these criticisms.

\textbf{Not the overall objective}. 
We recognized that we are far away from the actual reanalysis and forecasting goals of full state estimation. 
However, we argue that that is a rather ambitious challenge which will require a lot of interdisciplinary work across communities. 
In the meantime while we work towards that goal, operational centers could possibly improve their current products from ML-based techniques would would benefit downstream applications that deal directly with SSH.
Furthermore, SSH is an important variable in describing the full ocean state.
So a robust set of techniques that are able to solve the interpolation tasks could (in principal) be used to solve extra tasks.

\textbf{Small Region \& Period}.
We only feature a small region and period over the Gulfstream which is not representative of the different global regimes. 
This also does not take into account real things like \textit{data drift} which will inevitable occur in operational settings.
However, this is a dynamical regime and a well-studied area which does have some importance for specific communities and the results obtained offer some transferability to other dynamical regimes.
In addition, this area will have good coverage due to the new SWOT mission~\cite{SWOT} which will allow for further validation in the future.
Lastly, the area is small enough where the beginning stages for ML researchers is not overwhelmed with problems involving scale (even though we eventually want to arrive at global schemes).
We hope to extend our challenges to more relevant scenarios~\cite{MDSALONGTRACK}.

\textbf{Simulations versus Reanalysis}. We use simulations for the OSSE experiments instead of reanalysis. This is an open research question as it is unclear whether it's better to pretrain models on simulated ocean data or reanalysis ocean data. In future updates, we plan to add the reanalysis data to extend the challenge.

\textbf{Efficacy of OSSE Experiments}. We alluded to the idea that the OSSE experiments may not reflect the overarching goal of the user yet we provide more OSSE experiments than OSE experiments.
We acknowledged that it often does not coincide exactly with the OSE experiments which may give users a false sense of accomplishment and immediate transferability. 
However, we try to provide a framework where one could thoroughly experiment with the learning problem on OSSE configurations which can facilitate transfer learning to other domain-specific tasks.
We also anticipate that new \textit{real} SWOT data~\cite{SWOT} will start to become more available which will allow us to design better, realistic OSE experiments.

\textbf{Noise Characterization}.
Real data has noise to content with and we do not account for that within the SSH interpolation experiments.
The true noise we see in operational settings is structured and this would require more knowledge outside the scope of our teams expertise.
A more improved challenge would take these considerations into account.
We leave this as a future challenge for the community and we hope our platform can help facilitate this.

\textbf{Uncertainty Quantitification}.
We prefaced the problem statement with the idea of data assimilation which is the notion of \textit{state/parameter estimation under uncertain conditions and incomplete information}~\citep{DAGEOSCIENCE}.
However, we have not addressed any notion of uncertainty at all throughout the paper.
Uncertainty is difficult to quantify and we don't want to impose too many restrictions until we more sure about the efficacy of ML for easier problems.
However, to move the problem setting towards a more realistic setting, we can start to introduce metrics and additional requirements from future challenges, e.g. mean and standard deviation estimates or ensemble predictions.

\textbf{Operational Constraints}.
The real use case of SSH interpolation will involve global data and/or high-resolution data. 
This involves dealing with very high-dimensional spatiotemporal global state-space.
In practice, the necessity for the scalability of the method is of paramount importance.
However, there are also areas within the ML research community who are looking into many ways we can scale up physical models~\citep{VEROS,OCEANANIGANS} and machine learn models for geoscience tasks~\citep{SFNO}.
We anticipate that once a set of solutions are excepted by a community, the scalability will come later.

\end{document}